\documentclass[10pt, a4paper]{article}
\usepackage{lrec}
\usepackage{graphicx}
\usepackage{framed}
\usepackage{forest}
\usepackage{tabularx}
\usepackage{gensymb}
\usepackage{multirow}
\forestset{
  dir node/.style={
    parent anchor=south west,
    child anchor=west,
    anchor=west,
    inner ysep=-4pt,
    align=left,
    font=\itshape\footnotesize,
  },
  dir tree/.style={
    for tree={
      grow'=0,
      dir node,
      edge path={
        \noexpand\path [draw, \forestoption{edge}] (!u.parent anchor) ++(1em,-0.17em) |- node[fill,inner sep=1.25pt] {} (.child anchor)\forestoption{edge label};
      },
      if n children=0{}{
        font=\bfseries\small,
        delay={
          prepend={[text 1, dir node, phantom, calign with current]}
        }
      },
      fit=band,
      before computing xy={
        l=1.5em,
      }
    },
  }
}
\pgfdeclarelayer{background}
\pgfsetlayers{background,main}
\usepackage{tabularx}
\usepackage{soul}
\usepackage{subcaption}
\usepackage{epstopdf}
\usepackage[latin1]{inputenc}
\usepackage{hyperref}
\usepackage{xstring}

\title{A~Framework~for~Evaluation~of~Machine~Reading~Comprehension Gold~Standards} 

\name{Viktor Schlegel, Marco Valentino, Andr\'e Freitas, Goran Nenadic, Riza Batista-Navarro}

\address{Department of Computer Science, University of Manchester \\
         Manchester, United Kingdom \\
         \{viktor.schlegel, marco.valentino, andre.freitas, gnenadic, 	riza.batista\}@manchester.ac.uk\\}

\usepackage{xcolor}
\usepackage{tikz}
\usepackage{pgfplots}
\definecolor{bblue}{HTML}{4F81BD}
\definecolor{rred}{HTML}{C0504D}
\definecolor{ggreen}{HTML}{9BBB59}
\definecolor{ppurple}{HTML}{9F4C7C}

\usepackage{dirtree}
\usepackage{tabularx}
\usepackage{amsmath}
\pgfplotsset{
   compat=1.14,
   legend entry/.initial=,
   every axis plot post/.code={%
       \pgfkeysgetvalue{/pgfplots/legend entry}\tempValue
       \ifx\tempValue\empty
           \pgfkeysalso{/pgfplots/forget plot}%
       \else
           \expandafter\addlegendentry\expandafter{\tempValue}%
       \fi
   },
}

\abstract{
Machine Reading Comprehension (MRC) is the task of answering a question over a paragraph of text. While neural MRC systems gain popularity and achieve noticeable performance, issues are being raised with the methodology used to establish their performance, particularly concerning the data design of gold standards that are used to evaluate them. There is but a limited understanding of the challenges present in this data, which makes it hard to draw comparisons and formulate reliable hypotheses. 
As a first step towards alleviating the problem, this paper proposes a unifying framework to systematically investigate the present linguistic features, required reasoning and background knowledge and factual correctness on one hand, and the presence of lexical cues as a lower bound for the requirement of understanding on the other hand. We propose a qualitative annotation schema for the first and a set of approximative metrics for the latter.
In a first application of the framework, we analyse modern MRC gold standards and present our findings: the absence of features that contribute towards lexical ambiguity, the varying factual correctness of the expected answers and the presence of lexical cues, all of which potentially lower the reading comprehension complexity and quality of the evaluation data.  \\ \newline \Keywords{Machine Reading Comprehension, Question Answering, Evaluation Methodology, Annotation Schema} }

\begin{document}

\maketitleabstract

\section{Introduction}

There is a recent spark of interest in the task of Question Answering (QA) over unstructured textual data, also referred to as Machine Reading Comprehension (MRC).
This is mostly due to wide-spread success of advances in various facets of deep learning related research, such as novel architectures \cite{Vaswani2017,Sukhbaatar2015} that allow for efficient optimisation of neural networks consisting of multiple layers, hardware designed for deep learning purposes\footnote{https://cloud.google.com/tpu/}\footnote{https://www.nvidia.com/en-gb/data-center/tesla-v100/} and software frameworks \cite{abadi2016tensorflow,Paszke2017AutomaticPyTorch} that allow efficient development and testing of novel approaches. These factors enable researchers to produce models that are pre-trained on large scale corpora and provide contextualised word representations \cite{Peters2018} that are shown to be a vital component towards solutions for a variety of natural language understanding tasks, including MRC \cite{Devlin2018}. Another important factor that led to the recent success in MRC-related tasks is the widespread availability of various large datasets, e.g., SQuAD \cite{rajpurkar2016squad}, that provide sufficient examples for optimising statistical models. The combination of these factors yields notable results, even surpassing human performance \cite{Lan2020ALBERT:Representations}.     
\begin{figure}[!tb]
    \centering
\centering
\begin{tabularx}{1\columnwidth}{| X |}
\hline
\textbf{Passage 1: Marietta Air Force Station}\\
\emph{Marietta Air Force Station (ADC ID: M-111, NORAD ID: Z-111) is a 
closed United States Air Force General Surveillance Radar station. It is located 2.1 mi  northeast of Smyrna, Georgia. It was closed in 1968.}\\
\hline
\textbf{Passage 2: Smyrna, Georgia}\\
\emph{Smyrna is a city northwest of the neighborhoods of Atlanta. [\ldots]
As of the {\color{blue}2010} census, the city had a population of \textbf{51,271}. The U.S. Census Bureau estimated the population in 2013 to be 53,438. [\ldots]}\\
\hline
\hline
\textbf{Question:} \emph{What is the {\color{blue}2010} population of the city 2.1 miles southwest of Marietta Air Force Station?} \\
\hline
\end{tabularx}
    \caption{While initially this looks like a complex question that requires the synthesis of different information across multiple documents, the keyword ``2010'' appears in the question and only in the sentence that answers it, considerably simplifying the search. Full example with 10 passages can be seen in Appendix~\ref{sec:appendix-example}.}
    \label{fig:exploitable-example}
\end{figure}

MRC is a generic task format that can be used to probe for various natural language understanding capabilities \cite{gardner2019question}. Therefore it is crucially important to establish a rigorous evaluation methodology in order to be able to draw reliable conclusions from conducted experiments. While increasing effort is put into the evaluation of novel architectures, such as keeping the evaluation data from public access to prevent unintentional overfitting to test data, performing ablation and error studies and introducing novel metrics \cite{dodge2019show}, surprisingly little is done to establish the quality of the data itself. Additionally, 
recent research arrived at worrisome findings: the data of those gold standards, which is usually gathered involving a crowd-sourcing step, suffers from flaws in design \cite{Chen2019a} or contains overly specific keywords \cite{Jia2017}.
Furthermore, these gold standards contain ``annotation artefacts'', cues that lead models into focusing on superficial aspects of text, such as lexical overlap and word order, instead of actual language understanding \cite{mccoy2019right,gururangan2018annotation}. These weaknesses cast some doubt on whether the data can reliably evaluate the \emph{reading} comprehension performance of the models they evaluate, i.e. if the models are indeed being assessed for their capability to read. 

Figure~\ref{fig:exploitable-example} shows an example from \textsc{HotpotQA} \cite{Yang2018}, a dataset that exhibits the last kind of weakness mentioned above, i.e., the presence of unique keywords in both the question and the passage (in close proximity to the expected answer).

An evaluation methodology is vital to the fine-grained understanding of challenges associated with a single gold standard, in order to understand in greater detail which capabilities of MRC models it evaluates. More importantly, it allows to draw comparisons between multiple gold standards and between the results of respective state-of-the-art models that are evaluated on them. 

In this work, we take a step back and propose a framework to systematically analyse MRC evaluation data, typically a set of questions and expected answers to be derived from accompanying passages. Concretely, we introduce a methodology to categorise the \emph{linguistic complexity} of the textual data and the \emph{reasoning} and potential external \emph{knowledge} required to obtain the expected answer. Additionally we propose to take a closer look at the \emph{factual correctness} of the expected answers, a quality dimension that appears under-explored in literature.

We demonstrate the usefulness of the proposed framework by applying it to precisely describe and compare six contemporary MRC datasets. Our findings reveal concerns about their factual correctness, the presence of lexical cues that simplify the task of reading comprehension and the lack of semantic altering grammatical modifiers. We release the raw data comprised of 300 paragraphs, questions and answers richly annotated under the proposed framework as a resource for researchers developing natural language understanding models and datasets to utilise further. 
 
To the best of our knowledge this is the first attempt to introduce a common evaluation methodology for MRC gold standards and the first across-the-board qualitative evaluation of  MRC datasets with respect to the proposed categories.

\section{Framework for MRC Gold Standard Analysis}

\subsection{Problem definition}
\label{sec:problem-definition}
We define the task of machine reading comprehension, the target application of the proposed methodology as follows: Given a paragraph $P$ that consists of tokens (words) $p_1, \ldots, p_{n_P}$ and a question $Q$ that consists of tokens $q_1 \ldots q_{n_Q}$, the goal is to retrieve an answer $A$ with tokens $a_1 \ldots a_{n_A}$. $A$ is commonly constrained to be one of the following cases \cite{Liu2019a}, illustrated in Figure~\ref{table:qa-formulation}:
\begin{itemize}
    \item \textbf{Multiple choice}, where the goal is to predict $A$ from a given set of choices $\mathcal{A}$. 
    \item \textbf{Cloze-style}, where $S$ is a sentence, and $A$ and $Q$ are obtained by removing a sequence of words such that $Q = S - A$. The task is to fill in the resulting gap in $Q$ with the expected answer $A$ to form $S$.
    \item \textbf{Span}, where is a continuous subsequence of tokens from the paragraph ($A \subseteq P$).  Flavours include multiple spans as the correct answer or $A \subseteq Q$.
    \item \textbf{Free form}, where $A$ is an unconstrained natural language string.
\end{itemize}

\begin{figure}[!tb]
\centering
\begin{tabularx}{1\columnwidth}{| X |}
\hline
\textbf{Passage}\\
\emph{The Pats win the AFC East for the 9th straight year. The Patriots trailed 24-16 at the end of the third quarter. They scored on a 46-yard field goal with 4:00 left in the game to pull within 24-19. Then, with 56 seconds remaining, Dion Lewis scored on an 8-yard run and the Patriots added a two-point conversion to go ahead 27-24. [\ldots] The game ended on a Roethlisberger interception. Steelers wide receiver Antonio Brown left in the first half with a bruised calf.}\\
\hline
\hline
\textbf{Multiple choice} \\
\emph{Question: Who was injured during the match?} \\
\emph{Answer: (a) Rob Gronkowski (b) Ben Roethlisberger (c) Dion Lewis \textbf{(d) Antonio Brown}}\\
\hline
\hline
\textbf{Cloze-style} \\
\emph{Question: The Patriots champion the cup for $\star$ consecutive seasons.} \\
\emph{Answer: \textbf{9}} \\
\hline
\hline
\textbf{Span} \\
\emph{Question: What was the final score of the game?} \\
\emph{Answer: \textbf{27-24}} \\
\hline
\hline
\textbf{Free form} \\
\emph{Question: How many points ahead were the Patriots by the end of the game?} \\
\emph{Answer: \textbf{3}} \\
\hline
\end{tabularx}
\caption{Typical formulations of the MRC task}
\label{table:qa-formulation}
\end{figure}

A gold standard $G$ is composed of $m$ entries $(Q_i, A_i, P_i)_{i\in\{1,\ldots, m\}}$.

The performance of an approach is established by comparing its answer predictions $A^*_{i}$ on the given input $(Q_i, T_i)$ (and $\mathcal{A}_i$ for the multiple choice setting) against the expected answer $A_i$ for all $i\in\{1,\ldots, m\}$ under a performance metric. Typical performance metrics are \emph{exact match (EM)} or \emph{accuracy}, i.e. the percentage of exactly predicted answers, and the \emph{F1 score} -- the harmonic mean between the precision and the recall of the predicted tokens compared to expected answer tokens. The overall F1 score can either be computed by averaging the F1 scores for every instance or by first averaging the precision and recall and then computing the F1 score from those averages (macro F1). Free-text answers, meanwhile, are evaluated by means of text generation and summarisation metrics such as BLEU \cite{Papineni2001} or ROUGE-L \cite{Lin2004Rouge:Summaries}.

\begin{figure}[!htb]
\begin{framed}
\begin{forest}
  dir tree,
  [Annotation Schema
    [Supporting Fact]
    [Answer Type
        [Paraphrasing - Generated\\
        Span - Unanswerable]
    ]
    [Correctness
        [Debatable - Wrong]
    ]
    [Reasoning
        [Operational
            [Bridge - Comparison - Constraint - Intersection]
        ]
        [Arithmetic
            [Substraction - Addition\\
             Ordering - Counting - Other]
        ]
        [Linguistic
            [Negation - Quantifiers - Conditional\\
            Monotonicity - Con-/Disjunction 
            ]
        ]
        [Temporal]
        [Spatial]
        [Causal]
        [By Exclusion]
        [Retrieval]
    ]
    [Knowledge
        [Factual,
            [Cultural/Historic - (Geo)Political/Legal\\
            Technical/Scientific - Other Domain Specific]
        ]
        [Intuitive]
    ]
    [Linguistic Complexity
        [Lexical Variety
            [Redundancy - Lexical Entailment\\
            Dative - Synonym/Paraphrase\\
            Abbreviation - Symmetry]
        ]
        [Syntactic Variety
            [Nominalisation - Genitive - Voice]
        ]
        [Lexical Ambiguity
            [Restrictivity - Factivity\\
            Coreference - Ellipse/Implicit]
        ]
        [Syntactic Ambiguity
            [Preposition - Listing - Coordination Scope\\
            Relative Clause/Adverbial/Apposition
            ]
        ]
    ]
  ]
\end{forest}
\end{framed}
\caption{The hierarchy of categories in our proposed annotation framework. Abstract higher-level categories are presented in bold while actual annotation features are shown in italics.}
\label{fig:taxonomy}
\end{figure}
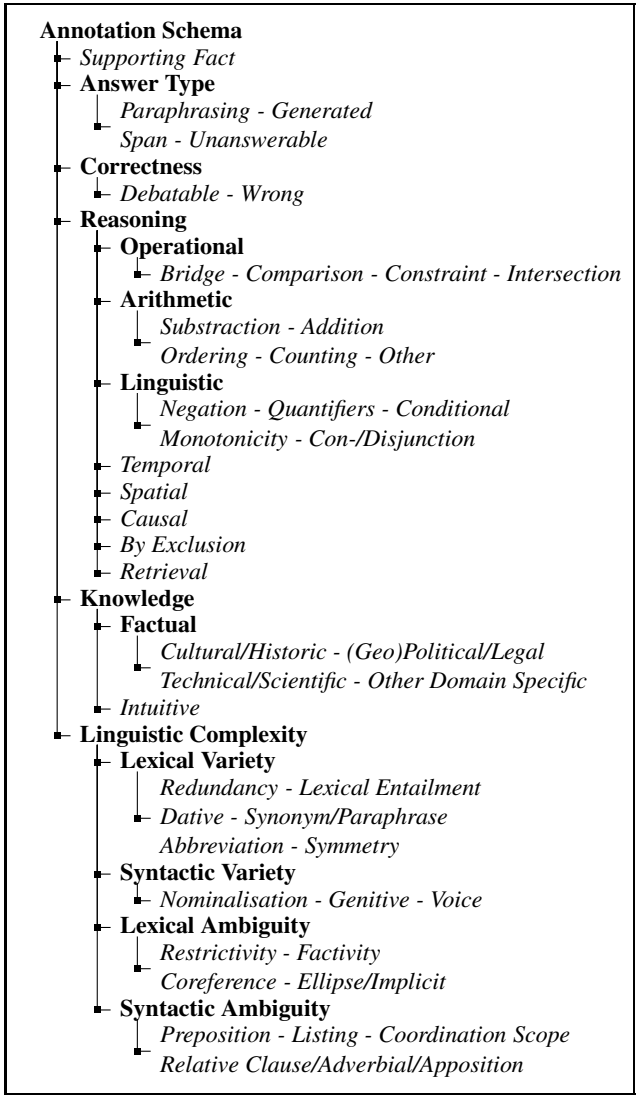

\subsection{Dimensions of Interest}
\label{sec:dimensions}
In this section we describe a methodology to categorise gold standards according to linguistic complexity, required reasoning and background knowledge, and their factual correctness. Specifically, we use those dimensions as high-level categories of a qualitative annotation schema for annotating question, expected answer and the corresponding context. We further enrich the qualitative annotations by a metric based on lexical cues in order to approximate a lower bound for the complexity of the reading comprehension task. 
By sampling entries from each gold standard and annotating them, we obtain measurable results and thus are able to make observations about the challenges present in that gold standard data. 

\paragraph{Problem setting}
We are interested in different types of the expected answer. We differentiate between \emph{Span}, where an answer is a continuous span taken from the passage, \emph{Paraphrasing}, where the answer is a paraphrase of a text span, \emph{Unanswerable}, where there is no answer present in the context, and \emph{Generated}, if it does not fall into any of the other categories. It is not sufficient for an answer to restate the question or combine multiple \emph{Span} or \emph{Paraphrasing} answers to be annotated as \emph{Generated}. It is worth mentioning that we focus our investigations on answerable questions. For a complementary qualitative analysis that categorises unanswerable questions, the reader is referred to \newcite{Yatskar2019}.

Furthermore, we mark a sentence as \emph{Supporting Fact} if it contains evidence required to produce the expected answer, as they are used further in the complexity analysis.

\paragraph{Factual Correctness} An important factor for the quality of a benchmark is its factual correctness, because on the one hand, the presence of factually wrong or debatable examples introduces an upper bound for the achievable performance of models on those gold standards. On the other hand, it is hard to draw conclusions about the correctness of answers produced by a model that is evaluated on partially incorrect data. 

One way by which developers of modern crowd-sourced gold standards ensure quality is by having the same entry annotated by multiple workers \cite{Trischler2017} and keeping only those with high agreement. We investigate whether this method is enough to establish a sound ground truth answer that is unambiguously correct. Concretely we annotate an answer as \emph{Debatable} when the passage features multiple plausible answers, when multiple expected answers contradict each other, or an answer is not specific enough with respect to the question and a more specific answer is present. We annotate an answer as \emph{Wrong} when it is factually wrong and a correct answer is present in the context.

\paragraph{Required Reasoning} It is important to understand what types of reasoning the benchmark evaluates, in order to be able to accredit various reasoning capabilities to the models it evaluates. Our proposed reasoning categories are inspired by those found in scientific question answering literature \cite{Jansen2016,Boratko2018ADataset}, as research in this area focuses on understanding the required reasoning capabilities.
We include reasoning about the \emph{Temporal} succession of events,  \emph{Spatial} reasoning about directions and environment, and \emph{Causal} reasoning about the cause-effect relationship between events. We further annotate (multiple-choice) answers that can only be answered \emph{By Exclusion} of every other alternative.

We further extend the reasoning categories by operational logic, similar to those required in semantic parsing tasks \cite{Berant2013}, as solving those tasks typically requires ``multi-hop'' reasoning \cite{Yang2018,welbl2018constructing}. When an answer can only be obtained by combining information from different sentences joined by mentioning a common entity, concept, date, fact or event (from here on called entity), we annotate it as \emph{Bridge}. We further annotate the cases, when the answer is a concrete entity that satisfies a \emph{Constraint} specified in the question, when it is required to draw a \emph{Comparison} of multiple entities' properties or when the expected answer is an \emph{Intersection} of their properties (e.g. ``What do Person A and Person B have in common?'')

We are interested in the linguistic reasoning capabilities probed by a gold standard, therefore we include the appropriate category used by \newcite{Wang2019}. Specifically, we annotate occurrences that require understanding of \emph{Negation}, \emph{Quantifiers} (such as ``every'', ``some'', or ``all''), \emph{Conditional} (``if \ldots then'') statements and the logical implications of \emph{Con-/Disjunction} (i.e. ``and'' and ``or'') in order to derive the expected answer.

Finally, we investigate whether arithmetic reasoning requirements emerge in MRC gold standards as this can probe for reasoning that is not evaluated by simple answer retrieval \cite{Dua2019}. 
To this end, we annotate the presence of of \emph{Addition} and \emph{Subtraction}, answers that require \emph{Ordering} of numerical values, \emph{Counting} and \emph{Other} occurrences of simple mathematical operations.

An example can exhibit multiple forms of reasoning. Notably, we do not annotate any of the categories mentioned above if the expected answer is directly stated in the passage. For example, if the question asks ``How many total points were scored in the game?'' and the passage contains a sentence similar to ``The total score of the game was 51 points'', it does not require any reasoning, in which case we annotate it as \emph{Retrieval}.

\paragraph{Knowledge} Worthwhile knowing is whether the information presented in the context is sufficient to answer the question, as there is an increase of benchmarks deliberately designed to probe a model's reliance on some sort of background knowledge \cite{Storks2019}. We seek to categorise the type of knowledge required. 
Similar to \newcite{Wang2019}, on the one hand we annotate the reliance on factual knowledge, that is \emph{(Geo)political/Legal, Cultural/Historic, Technical/Scientific} and \emph{Other Domain Specific} knowledge about the world that can be expressed as a set of facts. On the other hand, we denote \emph{Intuitive} knowledge requirements, which is challenging to express as a set of facts, such as the knowledge that a parenthetic numerical expression next to a person's name in a biography usually denotes his life span.

\paragraph{Linguistic Complexity} Another dimension of interest is the evaluation of various linguistic capabilities of MRC models  \cite{Goldberg2019,Liu2019LinguisticRepresentations,Tenney2019BERTPipeline}. We aim to establish which linguistic phenomena are probed by gold standards and to which degree. To that end, we draw inspiration from the annotation schema used by \newcite{Wang2019}, and adapt it around lexical semantics and syntax.

More specifically, we annotate features that introduce variance between the supporting facts and the question. With regard to lexical semantics, we focus on the use of redundant words  that do not alter the meaning of a sentence for the task of retrieving the expected answer (\emph{Redundancy}), requirements on the understanding of words' semantic fields (\emph{Lexical Entailment})
and the use of \emph{Synonyms and Paraphrases} with respect to the question wording. Furthermore we annotate cases where supporting facts contain \emph{Abbreviations} of concepts introduced in the question (and vice versa) and when a \emph{Dative} case substitutes the use of a preposition (e.g. ``I bought her a gift'' vs ``I bought a gift for her''). Regarding syntax, we annotate changes from passive to active \emph{Voice}, the substitution of a \emph{Genitive} case with a preposition (e.g. ``of'') and changes from nominal to verbal style and vice versa (\emph{Nominalisation}). 

We recognise features that add ambiguity to the supporting facts, for example when information is only expressed implicitly by using an \emph{Ellipsis}. As opposed to redundant words, we annotate \emph{Restrictivity} and \emph{Factivity} modifiers, words and phrases whose presence does change the meaning of a sentence with regard to the expected answer, and occurrences of intra- or inter-sentence \emph{Coreference} in supporting facts (that is relevant to the question). Lastly, we mark ambiguous syntactic features, when their resolution is required in order to obtain the answer. Concretely, we mark  argument collection with con- and disjunctions (\emph{Listing}) and ambiguous \emph{Prepositions}, \emph{Coordination Scope} and \emph{Relative clauses/Adverbial phrases/Appositions}.


\paragraph{Complexity} Finally, we want to approximate the presence of lexical cues that might simplify the reading required in order to arrive at the answer. Quantifying this allows for more reliable statements about and comparison of the complexity of gold standards, particularly regarding the evaluation of comprehension that goes beyond simple lexical matching. We propose the use of coarse metrics based on lexical overlap between question and context sentences.
Intuitively, we aim to quantify how much supporting facts ``stand out'' from their surrounding passage context. This can be used as proxy for the capability to retrieve the answer \cite{Chen2019a}.
Specifically, we measure \emph{(i)} the number of words jointly occurring in a question and a sentence, \emph{(ii)} the length of the longest n-gram shared by question and sentence and \emph{(iii)} whether a word or n-gram from the question uniquely appears in a sentence.

The resulting taxonomy of the framework is shown in Figure~\ref{fig:taxonomy}. The full catalogue of features, their description, detailed annotation guideline as well as illustrating examples can be found in Appendix~\ref{sec:appendix-schema}.


\section{Application of the Framework}
\subsection{Candidate Datasets}
We select contemporary MRC benchmarks to represent all four commonly used problem definitions \cite{Liu2019a}. In selecting relevant datasets, we do not consider those that are considered ``solved'', i.e. where the state of the art performance surpasses human performance, as is the case with \textsc{SQuAD} \cite{rajpurkar2018know,Lan2020ALBERT:Representations}. Concretely, we selected gold standards that fit our problem definition and were published in the years 2016 to 2019, have at least $(2019 - publication\ year) \times 20$ citations,
and bucket them according to the answer selection styles as described in Section~\ref{sec:problem-definition} We randomly draw one from each bucket and add two randomly drawn datasets from the candidate pool. 
This leaves us with the datasets described in Table~\ref{table:selected-datasets}. For a more detailed  description, we refer to Appendix~\ref{sec:appendix-datasets}.

\begin{table}[!tb]
\centering
\begin{tabularx}{0.95\columnwidth}{|l  l  X|}
\hline
\multicolumn{3}{|l|}{\textbf{Dataset}} \\
\textbf{\# passages} & \textbf{\# questions} & \textbf{Style} \\
\hline
\multicolumn{3}{|l|}{\textsc{MSMarco} \cite{nguyen2016ms}} \\
101093 & 101093  &   Free Form \\
\hline
\multicolumn{3}{|l|}{\textsc{HotpotQA} \cite{Yang2018}} \\
7405 & 7405 & Span, Yes/No \\
\hline
\multicolumn{3}{|l|}{\textsc{ReCoRd} \cite{Zhang2018}} \\
7279 & 10000 & Cloze-Style \\
\hline
\multicolumn{3}{|l|}{\textsc{MultiRC} \cite{Khashabi2018}} \\
81 & 953 & Multiple Choice \\
\hline
\multicolumn{3}{|l|}{\textsc{NewsQA} \cite{Trischler2017}} \\
637 & 637 & Span \\
\hline
\multicolumn{3}{|l|}{\textsc{DROP} \cite{Dua2019}} \\
588 & 9622 & Span, Numbers \\
\hline
\end{tabularx}
\caption{Summary of selected datasets}
\label{table:selected-datasets}
\end{table}

\subsection{Annotation Task}
We randomly select 50 distinct question, answer and passage triples from the publicly available development sets of the described datasets. Training, development and the (hidden) test set are drawn from the same distribution defined by the data collection method of the respective dataset. For those collections that contain multiple questions over a single passage, we ensure that we are sampling unique paragraphs in order to increase the variety of investigated texts. 

The samples were annotated by the first author of this paper, using the proposed schema. 
In order to validate our findings, we further take 20\% of the annotated samples and present them to a second annotator (second author). Since at its core, the annotation is a multi-label task, we report the inter-annotator agreement by computing the (micro-averaged) F1 score, where we treat the first annotator's labels as gold. Table~\ref{tab:annotator-agreement-per-category} reports the agreement scores, the overall (micro) average F1 score of the annotations is 0.82, which means that on average, more than two thirds of the overall annotated labels were agreed on by both annotators. We deem this satisfactory, given the complexity of the annotation schema.

\begin{table}[!b]
\centering
\begin{tabularx}{1\columnwidth}{| X | c |}
\hline
\textbf{Dataset} & \textbf{F1 Score} \\
\hline
\textsc{MsMarco} & 0.86 \\
\textsc{HotpotQA} & 0.88 \\
\textsc{ReCoRd} & 0.73 \\
\textsc{MultiRC} & 0.75 \\
\textsc{NewsQA} & 0.87 \\
\textsc{DROP} & 0.85\\
\hline
\hline
Micro Average & 0.82\\
\hline
\end{tabularx}
\caption{Inter-Annotator agreement F1 scores, averaged for each dataset}
\label{tab:annotator-agreement-per-category}
\end{table}


\subsection{Qualitative Analysis}
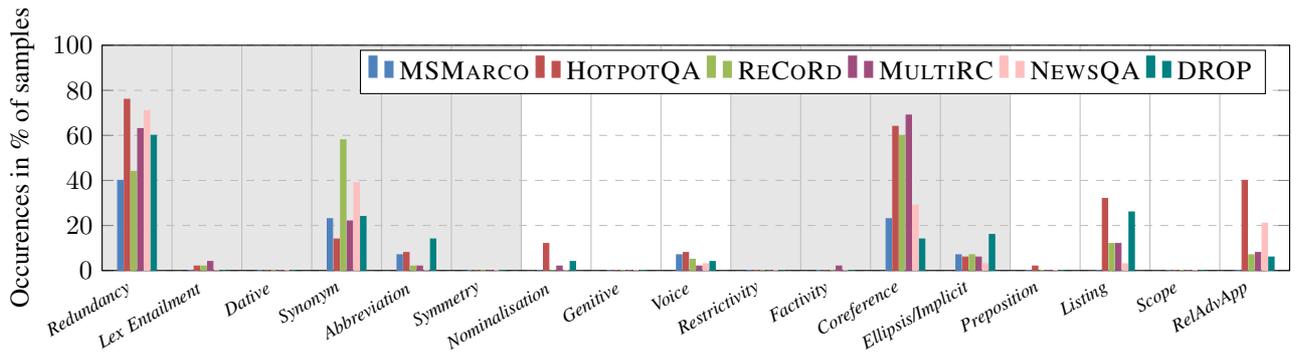
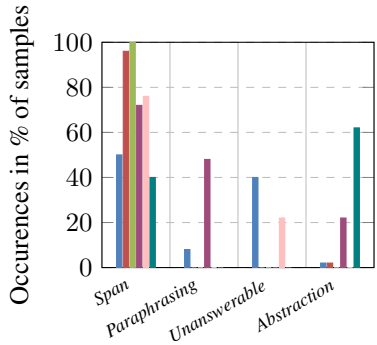
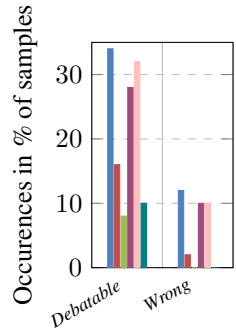
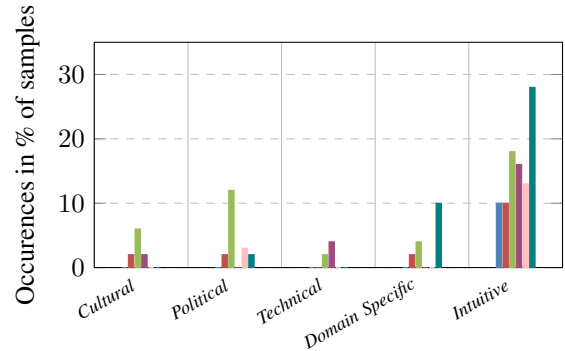
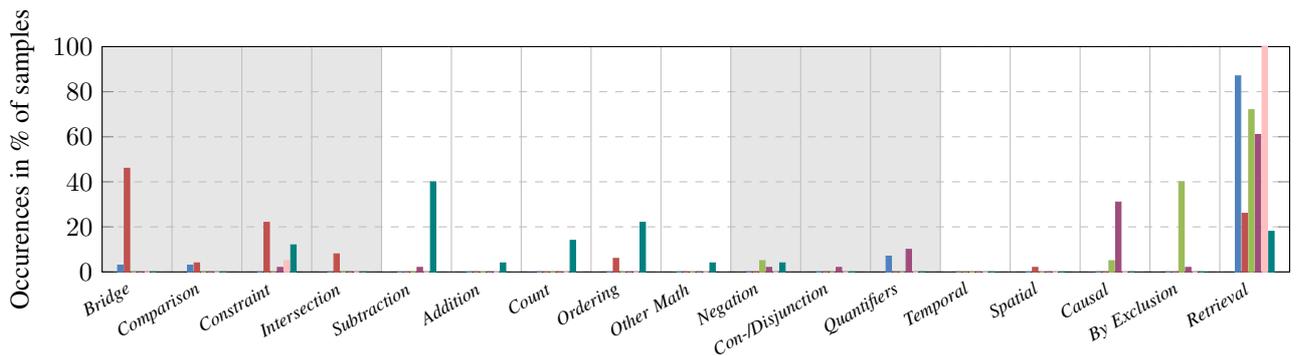
\begin{figure*}[t]
    \centering
    
\begin{subfigure}[t]{1\textwidth}
    \centering
    \begin{tikzpicture}
\begin{axis}[
height=13em,
    width  = \textwidth,
    ybar=0.5pt,
    ymin=0,
    ymax=100,
    scaled y ticks = false,
    bar width=2pt,
    x tick label style={rotate=30, anchor=east, align=right,text width=2cm, font=\scriptsize\itshape},
    major x tick style = transparent,
    ylabel={Occurences in \% of samples},
    xmajorgrids=true,
    x tick label as interval,
    ymajorgrids=true,
    y grid style=dashed,
    legend columns=-1,
    xticklabels={Redundancy,Lex Entailment,Dative,Synonym,Abbreviation,Symmetry,Nomi\-nalisation,Genitive,Voice,Restrictivity,Factivity,Coreference,Ellipsis/Implicit,Preposition,Listing,Scope,RelAdvApp},
    xtick={0,1,2,3,4,5,6,7,8,9,10,11,12,13,14,15,16,17},
    xmin=0, xmax=17,
        ]
    \addplot+[legend entry=\textsc{MSMarco}, color=bblue, fill=bblue] 
coordinates {
    (0.5,40) (1.5,0) (2.5,0) (3.5,23) (4.5,7) (5.5,0) (6.5,0) (7.5,0) (8.5,7) (9.5,0) (10.5,0) (11.5,23) (12.5,7) (13.5,0) (14.5,0) (15.5,0) (16.5,0)
    };
\addplot+[legend entry=\textsc{HotpotQA}, color=rred, fill=rred] 
coordinates {
    (0.5,76) (1.5,2) (2.5,0) (3.5,14) (4.5,8) (5.5,0) (6.5,12) (7.5,0) (8.5,8) (9.5,0) (10.5,0) (11.5,64) (12.5,6) (13.5,2) (14.5,32) (15.5,0) (16.5,40)
    };
\addplot+[legend entry=\textsc{ReCoRd}, color=ggreen, fill=ggreen] 
coordinates {
    (0.5,44) (1.5,2) (2.5,0) (3.5,58) (4.5,2) (5.5,0) (6.5,0) (7.5,0) (8.5,5) (9.5,0) (10.5,0) (11.5,60) (12.5,7) (13.5,0) (14.5,12) (15.5,0) (16.5,7)
    };
\addplot+[legend entry=\textsc{MultiRC}, color=ppurple, fill=ppurple] 
coordinates {
    (0.5,63) (1.5,4) (2.5,0) (3.5,22) (4.5,2) (5.5,0) (6.5,2) (7.5,0) (8.5,2) (9.5,0) (10.5,2) (11.5,69) (12.5,6) (13.5,0) (14.5,12) (15.5,0) (16.5,8)
    };
\addplot+[legend entry=\textsc{NewsQA}, color=pink, fill=pink] 
coordinates {
    (0.5,71) (1.5,0) (2.5,0) (3.5,39) (4.5,0) (5.5,0) (6.5,0) (7.5,0) (8.5,3) (9.5,0) (10.5,0) (11.5,29) (12.5,3) (13.5,0) (14.5,3) (15.5,0) (16.5,21)
    };
\addplot+[legend entry=\textsc{DROP}, color=teal, fill=teal] 
coordinates {
    (0.5,60) (1.5,0) (2.5,0) (3.5,24) (4.5,14) (5.5,0) (6.5,4) (7.5,0) (8.5,4) (9.5,0) (10.5,0) (11.5,14) (12.5,16) (13.5,0) (14.5,26) (15.5,0) (16.5,6)
    };
\begin{pgfonlayer}{background}
  \fill[color=black!10] (axis cs:0,0) rectangle (axis cs:6,100);
\end{pgfonlayer}
\begin{pgfonlayer}{background}
  \fill[color=black!10] (axis cs:9,0) rectangle (axis cs:13,100);
\end{pgfonlayer}
\end{axis}
\end{tikzpicture}
    \vspace{-1.8\baselineskip}
    \caption{Lexical (grey background) and syntactic (white background) linguistic features}
    \label{fig:linguistic-features}
\end{subfigure}    
\begin{subfigure}[t]{0.3\textwidth}
    \begin{tikzpicture}
\begin{axis}[
    height=13em,
    width  = 1\textwidth,
    ybar=0.5pt,
    ymin=0,
    ymax=100,
    scaled y ticks = false,
    bar width=2pt,
    x tick label style={rotate=30, anchor=east, align=right,text width=2cm, font=\scriptsize\itshape},
    major x tick style = transparent,
    ylabel={Occurences in \% of samples},
    x tick label as interval,
    xmajorgrids=true,
    ymajorgrids=true,
    y grid style=dashed,
        xticklabels={Span,Para\-phrasing,Unans\-werable,Abstraction},
    xtick={0,1,2,3,4},
    xmin=0, xmax=4,
    ]
    \addplot+[bar shift=-6.25pt, color=bblue, fill=bblue] 
coordinates {
    (0.5,50) (1.5,8) (2.5,40) (3.5,2)
    };
\addplot+[bar shift=-3.75pt, color=rred, fill=rred] 
coordinates {
    (0.5,96) (1.5,0) (2.5,0) (3.5,2)
    };
\addplot+[bar shift=-1.25pt, color=ggreen, fill=ggreen] 
coordinates {
    (0.5,100) (1.5,0) (2.5,0) (3.5,0)
    };
\addplot+[bar shift=1.25pt, color=ppurple, fill=ppurple] 
coordinates {
    (0.5,72) (1.5,48) (2.5,0) (3.5,22)
    };
\addplot+[bar shift=3.75pt, color=pink, fill=pink] 
coordinates {
    (0.5,76) (1.5,0) (2.5,22) (3.5,0)
    };
\addplot+[bar shift=6.25pt, color=teal, fill=teal] 
coordinates {
    (0.5,40) (1.5,0) (2.5,0) (3.5,62)
    };
 
\end{axis}
\end{tikzpicture}
    \vspace{-1.8\baselineskip}
    \caption{Answer Type}
    
    \label{fig:answer-type-distribution}
\end{subfigure}
\begin{subfigure}[t]{0.2\textwidth}
    \centering
    \begin{tikzpicture}
\begin{axis}[
    height=13em,
    width=\textwidth,
    ybar=2*\pgflinewidth,
    ymin=0,
    ymax=35,
    scaled y ticks = false,
    bar width=2pt,
    x tick label style={rotate=30, anchor=east, align=right,text width=2cm, font=\scriptsize\itshape},
    major x tick style = transparent,
    ylabel={Occurences in \% of samples},
    x tick label as interval,
    xmajorgrids=true,
    ymajorgrids=true,
    y grid style=dashed,
    xticklabels={Debatable,Wrong},
    xtick={0,1,2},
    xmin=0, xmax=2,
    ]
    \addplot+[bar shift=-6.25pt, color=bblue, fill=bblue] 
coordinates {
    (0.5,34) (1.5,12)
    };
\addplot+[bar shift=-3.75pt, color=rred, fill=rred] 
coordinates {
    (0.5,16) (1.5,2)
    };
\addplot+[bar shift=-1.25pt, color=ggreen, fill=ggreen] 
coordinates {
    (0.5,8) (1.5,0)
    };
\addplot+[bar shift=1.25pt, color=ppurple, fill=ppurple] 
coordinates {
    (0.5,28) (1.5,10)
    };
\addplot+[bar shift=3.75pt, color=pink, fill=pink] 
coordinates {
    (0.5,32) (1.5,10)
    };
\addplot+[bar shift=6.25pt, color=teal, fill=teal] 
coordinates {
    (0.5,10) (1.5,0)
    };

\end{axis}
\end{tikzpicture}
    
    \vspace{-1.8\baselineskip}
    \caption{Factual Correctness}
    \label{fig:quality-distribution}
\end{subfigure}
\begin{subfigure}[t]{0.45\textwidth}
    \centering
    \begin{tikzpicture}
\begin{axis}[
    height=13em,
    width  = 1\textwidth,
    ybar=0.5pt,
    ymin=0,
    ymax=35,
    scaled y ticks = false,
    bar width=2pt,
    x tick label style={rotate=30, anchor=east, align=right,text width=2cm, font=\scriptsize\itshape},
    major x tick style = transparent,
    ylabel={Occurences in \% of samples},
    x tick label as interval,
    xmajorgrids=true,
    ymajorgrids=true,
    y grid style=dashed,
       xticklabels={Cultural,Political,Technical,Domain Specific,Intuitive},
    xtick={0,1,2,3,4,5},
    xmin=0, xmax=5,
    ]
    \addplot+[bar shift=-6.25pt, color=bblue, fill=bblue] 
coordinates {
    (0.5,0) (1.5,0) (2.5,0) (3.5,0) (4.5,10)
    };
\addplot+[bar shift=-3.75pt, color=rred, fill=rred] 
coordinates {
    (0.5,2) (1.5,2) (2.5,0) (3.5,2) (4.5,10)
    };
\addplot+[bar shift=-1.25pt, color=ggreen, fill=ggreen] 
coordinates {
    (0.5,6) (1.5,12) (2.5,2) (3.5,4) (4.5,18)
    };
\addplot+[bar shift=1.25pt, color=ppurple, fill=ppurple] 
coordinates {
    (0.5,2) (1.5,0) (2.5,4) (3.5,0) (4.5,16)
    };
\addplot+[bar shift=3.75pt, color=pink, fill=pink] 
coordinates {
    (0.5,0) (1.5,3) (2.5,0) (3.5,0) (4.5,13)
    };
\addplot+[bar shift=6.25pt, color=teal, fill=teal] 
coordinates {
    (0.5,0) (1.5,2) (2.5,0) (3.5,10) (4.5,28)
    };

\end{axis}
\end{tikzpicture}
    \vspace{-1.8\baselineskip}
    \caption{Required External Knowledge}
    \label{fig:background-distribution}
\end{subfigure}
\begin{subfigure}[t]{1\textwidth}
    \centering
        \begin{tikzpicture}
\begin{axis}[      
height=13em,
    width  = 1\textwidth,
    ybar=2*\pgflinewidth,
    ymin=0,
    ymax=100,
    scaled y ticks = false,
    bar width=2pt,
    x tick label style={rotate=30, anchor=east, align=right,text width=2cm, font=\scriptsize\itshape},
    major x tick style = transparent,
    ylabel={Occurences in \% of samples},
    x tick label as interval,
    xmajorgrids=true,
    ymajorgrids=true,
    y grid style=dashed,
    legend style={ anchor=north east },
    xticklabels={Bridge,Compa\-rison,Constraint,Inter\-section,Subtraction,Addition,Count,Ordering,Other Math,Negation,Con-/Disjunction,Quantifiers,Temporal,Spatial,Causal,By \mbox{Exclusion},Retrieval},
    xtick={0,1,2,3,4,5,6,7,8,9,10,11,12,13,14,15,16,17},
    xmin=0, xmax=17,
    ]
    \addplot+[bar shift=-6.25pt, color=bblue, fill=bblue] 
coordinates {
    (0.5,3) (1.5,3) (2.5,0) (3.5,0) (4.5,0) (5.5,0) (6.5,0) (7.5,0) (8.5,0) (9.5,0) (10.5,0) (11.5,7) (12.5,0) (13.5,0) (14.5,0) (15.5,0) (16.5,87)
    };
\addplot+[bar shift=-3.75pt, color=rred, fill=rred] 
coordinates {
    (0.5,46) (1.5,4) (2.5,22) (3.5,8) (4.5,0) (5.5,0) (6.5,0) (7.5,6) (8.5,0) (9.5,0) (10.5,0) (11.5,0) (12.5,0) (13.5,2) (14.5,0) (15.5,0) (16.5,26)
    };
\addplot+[bar shift=-1.25pt, color=ggreen, fill=ggreen] 
coordinates {
    (0.5,0) (1.5,0) (2.5,0) (3.5,0) (4.5,0) (5.5,0) (6.5,0) (7.5,0) (8.5,0) (9.5,5) (10.5,0) (11.5,0) (12.5,0) (13.5,0) (14.5,5) (15.5,40) (16.5,72)
    };
\addplot+[bar shift=1.25pt, color=ppurple, fill=ppurple] 
coordinates {
    (0.5,0) (1.5,0) (2.5,2) (3.5,0) (4.5,2) (5.5,0) (6.5,0) (7.5,0) (8.5,0) (9.5,2) (10.5,2) (11.5,10) (12.5,0) (13.5,0) (14.5,31) (15.5,2) (16.5,61)
    };
\addplot+[bar shift=3.75pt, color=pink, fill=pink] 
coordinates {
    (0.5,0) (1.5,0) (2.5,5) (3.5,0) (4.5,0) (5.5,0) (6.5,0) (7.5,0) (8.5,0) (9.5,0) (10.5,0) (11.5,0) (12.5,0) (13.5,0) (14.5,0) (15.5,0) (16.5,100)
    };
\addplot+[bar shift=6.25pt, color=teal, fill=teal] 
coordinates {
    (0.5,0) (1.5,0) (2.5,12) (3.5,0) (4.5,40) (5.5,4) (6.5,14) (7.5,22) (8.5,4) (9.5,4) (10.5,0) (11.5,0) (12.5,0) (13.5,0) (14.5,0) (15.5,0) (16.5,18)
    };
\begin{pgfonlayer}{background}
  \fill[color=black!10] (axis cs:0,0) rectangle (axis cs:4,100);
\end{pgfonlayer}
\begin{pgfonlayer}{background}
  \fill[color=black!10] (axis cs:9,0) rectangle (axis cs:12,100);
\end{pgfonlayer}
\end{axis}
\end{tikzpicture}
    \vspace{-1.8\baselineskip}
    \caption{Required operational, arithmetic and linguistic and other forms of Reasoning  (grouped from left to right)}
    \label{fig:operational-logic}
\end{subfigure}
\caption{Annotation results}
\label{fig:results}
\end{figure*}
We present a concise view of the annotation results in Figure~\ref{fig:results}. The full annotation results can be found in Appendix~\ref{sec:appendix-results}\footnote{Calculations and analysis code can be retrieved from \texttt{\url{https://github.com/schlevik/dataset-analysis}}}. We centre our discussion around the following main points:

\paragraph{Linguistic Features} As observed in Figure~\ref{fig:linguistic-features} the gold standards feature a high degree of \emph{Redundancy}, peaking at 76\% of the annotated \textsc{HotpotQA} samples and synonyms and paraphrases (labelled \emph{Synonym}), with \textsc{ReCoRd} samples containing 58\% of them, likely to be attributed to the elaborating type of discourse of the dataset sources (encyclopedia and newswire). 
This is, however, not surprising, as it is fairly well understood in the literature that current state-of-the-art models perform well on distinguishing relevant words and phrases from redundant ones \cite{Seo2017}.
Additionally, the representational capability of synonym relationships of word embeddings has been investigated and is well known \cite{chen2013expressive}. Finally, we observe the presence of syntactic features, such as ambiguous relative clauses, appositions and adverbial phrases, (\emph{RelAdvApp} 40\% in \textsc{HotpotQA} and ReCoRd) and those introducing variance, concretely switching between verbal and nominal styles (e.g. \emph{Nominalisation} 10\% in \textsc{HotpotQA}) and from passive to active voice (Voice, 8\% in \textsc{HotpotQA}). 

Syntactic features contributing to variety and ambiguity that we did not observe in our samples are the exploitation of verb symmetry, the use of dative and genitive cases or ambiguous prepositions and coordination scope (respectively \emph{Symmetry}, \emph{Dative}, \emph{Genitive}, \emph{Prepositions}, \emph{Scope}). Therefore we cannot establish whether models are capable of dealing with those features by evaluating them on those gold standards.

\begin{table}[!tb]
\centering
\begin{tabularx}{1\columnwidth}{| X  r|}
\hline
\textbf{Wrong Answer} & 25\%\\
\hline
\multicolumn{2}{|>{\hsize=\dimexpr2\hsize+2\tabcolsep+\arrayrulewidth\relax}X|}{\textbf{Question:} What is the cost of the project?}\\
\multicolumn{2}{|>{\hsize=\dimexpr2\hsize+2\tabcolsep+\arrayrulewidth\relax}X|}{\textbf{Expected Answer:} 2.9 Bio \$}\\
\multicolumn{2}{|>{\hsize=\dimexpr2\hsize+2\tabcolsep+\arrayrulewidth\relax}X|}{\textbf{Correct answer:} 4.1 Bio \$}\\
\multicolumn{2}{|>{\hsize=\dimexpr2\hsize+2\tabcolsep+\arrayrulewidth\relax}X|}{\textbf{Passage:} \emph{At issue is the alternate engine for the Joint Strike Fighter platform, [\ldots] that has cost taxpayers \$1.2 billion in earmarks since 2004. It is estimated to cost at least \$2.9 billion more until its completion.}} \\

\hline
\textbf{Answer Present}  & 47\%\\
\hline
\multicolumn{2}{|>{\hsize=\dimexpr2\hsize+2\tabcolsep+\arrayrulewidth\relax}X|}{\textbf{Question:} how long do you need to cook 6 pounds of pork in a roaster?} \\
\multicolumn{2}{|>{\hsize=\dimexpr2\hsize+2\tabcolsep+\arrayrulewidth\relax}X|}{\textbf{Expected Answer:} Unanswerable}\\
\multicolumn{2}{|>{\hsize=\dimexpr2\hsize+2\tabcolsep+\arrayrulewidth\relax}X|}{\textbf{Correct answer:} 150 min}\\
\multicolumn{2}{|>{\hsize=\dimexpr2\hsize+2\tabcolsep+\arrayrulewidth\relax}X|}{\textbf{Passage:} \emph{The rule of thumb for pork roasts is to cook them 25 minutes per pound of meat [\ldots]}} \\
\hline
\hline
\textbf{Arbitrary selection} & 25\%\\
\hline
\multicolumn{2}{|>{\hsize=\dimexpr2\hsize+2\tabcolsep+\arrayrulewidth\relax}X|}{\textbf{Question:} what did jolie say? }\\
\multicolumn{2}{|>{\hsize=\dimexpr2\hsize+2\tabcolsep+\arrayrulewidth\relax}X|}{\textbf{Expected Answer:} she feels passionate about Haiti}\\
\multicolumn{2}{|>{\hsize=\dimexpr2\hsize+2\tabcolsep+\arrayrulewidth\relax}X|}{\textbf{Passage:} \emph{Angelina Jolie says she feels passionate about Haiti, whose "extraordinary" people are inspiring her with their resilience after the devastating earthquake one month ago. During a visit to Haiti this week, she said that despite the terrible tragedy, Haitians are dignified and calm.}} \\
\hline
\textbf{Arbitrary Precision} & 33\%\\
\hline
\multicolumn{2}{|>{\hsize=\dimexpr2\hsize+2\tabcolsep+\arrayrulewidth\relax}X|}{\textbf{Question:} Where was the person killed Friday?} \\
\multicolumn{2}{|>{\hsize=\dimexpr2\hsize+2\tabcolsep+\arrayrulewidth\relax}X|}{\textbf{Expected Answer:} Arkansas }\\
\multicolumn{2}{|>{\hsize=\dimexpr2\hsize+2\tabcolsep+\arrayrulewidth\relax}X|}{\textbf{Passage:} \emph{The death toll from severe storms in northern Arkansas has been lowered to one person [\ldots].
Officials had initially said three people were killed when the storm and possible tornadoes walloped Van Buren County on Friday.}} \\
\hline
\end{tabularx}
\caption{Most frequently occurring factually wrong and debatable categories with an instantiating example. Percentages are relative to the number of all examples annotated as \emph{Wrong} respectively \emph{Debatable} across all six gold standards.}
\label{tab:examples}
\end{table}
\paragraph{Factual Correctness} We identify three common sources that surface in different problems regarding an answer's factual correctness, as reported in Figure \ref{fig:quality-distribution} and illustrate their instantiations in Table~\ref{tab:examples}: 
\begin{itemize}
    \item \textit{Design Constraints:} Choosing the task design and the data collection method introduces some constraints that lead to factually debatable examples. For example, a span might have been arbitrarily selected from multiple spans that potentially answer a question, but only a single continuous answer span per question is allowed by design, as observed in the \textsc{NewsQA} and \textsc{MsMarco} samples (32\% and 34\% examples annotated as \emph{Debatable} with 16\% and 53\% thereof exhibiting arbitrary selection, respectively). Sometimes, when additional passages are added after the annotation step, they can by chance contain passages that answer the question more precisely than the original span, as seen in \textsc{HotpotQA} (16\% \emph{Debatable} samples, 25\% of them due to arbitrary selection). In the case of \textsc{MultiRC} it appears to be inconsistent, whether multiple correct answer choices are expected to be correct in isolation or in conjunction (28\% \emph{Debatable} with 29\% of them exhibiting this problem). This might provide an explanation to its relatively weak human baseline performance of 84\% F1 score \cite{Khashabi2018}.
    \item \textit{Weak Quality assurance:} When the (typically crowd-sourced) annotations are not appropriately validated, incorrect examples will find their way into the gold standards.  This typically results in factually wrong expected answers (i.e. when a more correct answer is present in the context) or a question is expected to be Unanswerable, but is actually answerable from the provided context. The latter is observed in \textsc{MsMarco} (83\% of examples annotated as \emph{Wrong}) and \textsc{NewsQA}, where 60\% of the examples annotated as \emph{Wrong} are \emph{Unanswerable} with an answer present. 
    \item \textit{Arbitrary Precision:} There appears to be no clear guideline on how precise the answer is expected to be, when the passage expresses the answer in varying granularities. We annotated instances as \emph{Debatable} when the expected answer was not the most precise given the context (44\% and 29\% of \emph{Debatable} instances in \textsc{NewsQA} and \textsc{MultiRC}, respectively).
\end{itemize}


\paragraph{Semantics-altering grammatical modifiers} We took interest in whether any of the benchmarks contain what we call \textit{distracting lexical features} (or \textit{distractors}): grammatical modifiers that alter the semantics of a sentence for the final task of answering the given question while preserving a similar lexical form.
An example of such features are cues for (double) Negation (e.g., ``no'', ``not''), which when introduced in a sentence, reverse its meaning.
Other examples include modifiers denoting \emph{Restrictivity}, \emph{Factivity} and Reasoning (such as \emph{Monotonicity} and \emph{Conditional} cues).
Examples of question-answer pairs containing a distractor are shown in Table~\ref{tab:distractors}.

\begin{figure}[b]
    \centering
\begin{tabularx}{1\columnwidth}{| X |}
\hline
\textbf{Restrictivity Modification} \\
\hline
\textbf{Question:} What was the longest touchdown? \\
\textbf{Expected Answer:} 42 yard\\
\textbf{Passage:} \emph{Brady scored a 42 yard TD. Brady {\color{blue}almost} scored a 50 yard TD.} \\
\hline
\hline
\textbf{Factivity Altering} \\
\hline
\textbf{Question:} What are the details of the second plot on Alexander's life? \\
\textbf{(Wrong) Answer Choice:} Callisthenes of Olynthus was {\color{blue}definitely} involved.\\
\textbf{Passage:} \emph{[\ldots] His official historian, Callisthenes of Olynthus, was implicated in the plot; however, historians have yet to reach a consensus regarding this involvement.} \\
\hline
\hline
\textbf{Conditional Statement} \\
\hline
\textbf{Question:} How many eggs did I buy? \\
\textbf{Expected Answer:} 2.\\
\textbf{Passage:} \emph{[\ldots] I will buy 4 eggs, {\color{blue}if the market sells milk}. Otherwise, I will buy 2 [\ldots]. The market had no milk.} \\
\hline
\end{tabularx}
\caption{Example of semantics altering lexical features}
\label{tab:distractors}
\end{figure}
We posit that the presence of such distractors would allow for evaluating reading comprehension beyond potential simple word matching.
However, we observe no presence of such features in the benchmarks (beyond Negation in \textsc{DROP}, \textsc{ReCoRd} and \textsc{HotpotQA}, with 4\%, 4\% and 2\% respectively).
This results in gold standards that clearly express the evidence required to obtain the answer, lacking more challenging, i.e., distracting, sentences that can assess whether a model can truly understand meaning. 


\paragraph{Other} In the Figure~\ref{fig:operational-logic} we observe that \emph{Operational} and \emph{Arithmetic} reasoning moderately (6\% to 8\% combined) appears ``in the wild'', i.e. when not enforced by the data design as is the case with \textsc{HotpotQA} (80\% Operations combined) or \textsc{DROP} (68\% \emph{Arithmetic} combined). \emph{Causal} reasoning is (exclusively) present in \textsc{MultiRC} (32\%), whereas \emph{Temporal} and \emph{Spatial} reasoning requirements seem to not naturally emerge in gold standards. In \textsc{ReCoRd}, a fraction of 38\% questions can only be answered \emph{By Exclusion} of every other candidate, due to the design choice of allowing questions where the required information to answer them is not fully expressed in the accompanying paragraph.

Therefore, it is also a little surprising to observe that \textsc{ReCoRd} requires external resources with regard to knowledge, as seen in Figure~\ref{fig:background-distribution}. \textsc{MultiRC} requires technical or more precisely basic scientific knowledge (6\% \emph{Technical/Scientific}), as a portion of paragraphs is extracted from elementary school science textbooks \cite{Khashabi2018}.  Other benchmarks moderately probe for factual knowledge (0\% to 4\% across all categories), while \emph{Intuitive} knowledge is required to derive answers in each gold standard.

It is also worth pointing out, as done in Figure~\ref{fig:answer-type-distribution}, that although \textsc{MultiRC} and \textsc{MsMarco} are not modelled as a span selection problem, their samples  still contain 50\% and 66\% of answers that are directly taken from the context. \textsc{DROP} contains the biggest fraction of generated answers (60\%), due to the requirement of arithmetic operations.

To conclude our analysis, we observe similar distributions of linguistic features and reasoning patterns, except where there are constraints enforced by dataset design, annotation guidelines or source text choice. Furthermore, careful consideration of design choices (such as single-span answers) is required, to avoid impairing the factual correctness of datasets, as pure crowd-worker agreement seems not sufficient in multiple cases.



\subsection{Quantitative Results}

\paragraph{Lexical overlap}
We used the scores assigned by our proposed set of metrics (discussed in Section~\ref{sec:dimensions} Dimensions of Interest: Complexity) to predict the supporting facts in the gold standard samples (that we included in our manual annotation). Concretely, we used the following five features capturing lexical overlap: \emph{(i)} the number of words occurring in sentence and question, \emph{(ii)} the length of the longest n-gram shared by sentence and question, whether a \emph{(iii)} uni- and \emph{(iv)} bigram from the question is unique to a sentence, and \emph{(v)} the sentence index, as input to a logistic regression classifier. We optimised on each sample leaving one example for evaluation. We compute the average Precision, Recall and F1 score by means of leave-one-out validation with every sample entry. The averaged results after 5 runs are reported in Table~\ref{tab:logreg}.
\begin{table}[!tb]
\centering
\begin{tabularx}{1\columnwidth}{| X | c | c | c |}
\hline
\textbf{Dataset} & \textbf{P} & \textbf{R} & \textbf{F1}\\
\hline
\textsc{MsMarco} & 0.07 $\pm$.04 & 0.52 $\pm$.12 & 0.11 $\pm$.04\\
\textsc{HotpotQA} & 0.20 $\pm$.03 & 0.60 $\pm$.03 & 0.26 $\pm$.02\\
\textsc{ReCoRd} & 0.28 $\pm$.04 & 0.56 $\pm$.04 & 0.34 $\pm$.03\\
\textsc{MultiRC} & 0.37 $\pm$.04 & 0.59 $\pm$.05 & 0.40 $\pm$.03\\
\textsc{NewsQA} & 0.19 $\pm$.04 & 0.68 $\pm$.02 & 0.26 $\pm$.03\\
\textsc{DROP} & 0.62 $\pm$.02 & 0.80 $\pm$.01 & 0.66 $\pm$.02\\
\hline
\end{tabularx}
\caption{(Average) Precision, Recall and F1 score within the 95\% confidence interval of a linear classifier optimised on lexical features for the task of predicting supporting facts}
\label{tab:logreg}
\end{table}

We observe that even by using only our five features based lexical overlap, the simple logistic regression baseline is able to separate out the supporting facts from the context to a varying degree. This is in line with the lack of semantics-altering grammatical modifiers discussed in the qualitative analysis section above.
The classifier performs best on \textsc{DROP} (66\% F1) and \textsc{MultiRC} (40\% F1), which means that lexical cues can considerably facilitate the search for the answer in those gold standards.
On \textsc{MultiRC}, \newcite{yadav2019quick} come to a similar conclusion, by using a more sophisticated approach based on overlap between question, sentence and answer choices.

Surprisingly, the classifier is able to pick up a signal from supporting facts even on data that has been pruned against lexical overlap heuristics by populating the context with additional documents that have high overlap scores with the question. This results in significantly higher scores than when guessing randomly (\textsc{HotpotQA} 26\% F1, and \textsc{MsMarco} 11\% F1). We observe similar results in the case the length of the question leaves few candidates to compute overlap with $6.3$ and $7.3$ tokens on average for \textsc{MsMarco} and \textsc{NewsQA} (26\% F1), compared to $16.9$ tokens on average for the remaining four dataset samples. 


Finally, it is worth mentioning that although the queries in \textsc{ReCoRd} are explicitly independent from the passage, the linear classifier is still capable of achieving 34\% F1 score in predicting the supporting facts.

However, neural networks perform significantly better than our admittedly crude baseline (e.g. 66\% F1 for supporting facts classification on \textsc{HotpotQA} \cite{Yang2018}), albeit utilising more training examples, and a richer sentence representation. This facts implies that those neural models are capable of solving more challenging problems than simple ``text matching'' as performed by the logistic regression baseline. However, they still circumvent actual reading comprehension as the respective gold standards are of limited suitability to evaluate this \cite{Min2019,Jiang2019}. This suggests an exciting future research direction, that is categorising the scale between text matching and reading comprehension more precisely and respectively positioning state-of-the-art models thereon.





\section{Related Work}
Although not as prominent as the research on novel architecture, there has been steady progress in critically investigating the data and evaluation aspects of NLP and machine learning in general and MRC in particular.
\paragraph{Adversarial Evaluation} The authors of the \textsc{AddSent} algorithm \cite{Jia2017} show that MRC models trained and evaluated on the \textsc{SQuAD} dataset pay too little attention to details that might change the semantics of a sentence, and propose a crowd-sourcing based method to generate adversary examples to exploit that weakness. This method was further adapted to be fully automated \cite{Wang2018} and applied to different gold standards \cite{Jiang2019}. Our proposed approach differs in that we aim to provide qualitative justifications for those quantitatively measured issues.
\paragraph{Sanity Baselines} Another line of research establishes sane baselines to provide more meaningful context to the raw performance scores of evaluated models. When removing integral parts of the task formulation such as question, the textual passage or parts thereof \cite{Kaushik2019} or restricting model complexity by design in order to suppress some required form of reasoning \cite{chen2019understanding}, models are still able to perform comparably to the state-of-the-art. This raises concerns about the perceived benchmark complexity and is related to our work in a broader sense as one of our goals is to estimate the complexity of benchmarks.

\paragraph{Benchmark evaluation in NLP} Beyond MRC, efforts similar to ours that pursue the goal of analysing the evaluation of established datasets exist in Natural Language Inference \cite{gururangan2018annotation,mccoy2019right}. Their analyses reveal the existence of biases in training and evaluation data that can be approximated with simple majority-based heuristics. Because of these biases, trained models fail to extract the semantics that are required for the correct inference. Furthermore, a fair share of work was done to reveal gender bias in coreference resolution datasets and models \cite{Rudinger2018GenderResolution,Zhao2018GenderMethods,Webster2018}.

\paragraph{Annotation Taxonomies} Finally, related to our framework are works that introduce annotation categories for gold standards evaluation. Concretely, we build our annotation framework around linguistic features that were introduced in the \textsc{GLUE} suite \cite{Wang2019} and the reasoning categories introduced in the \textsc{WorldTree} dataset \cite{Jansen2016}. A qualitative analysis complementary to ours, with focus on the unanswerability patterns in datasets that feature unanswerable questions was done by \newcite{Yatskar2019}.



\section{Conclusion}
In this paper, we introduce a novel framework to characterise machine reading comprehension gold standards. This framework has potential applications when comparing different gold standards, considering the design choices for a new gold standard and performing qualitative error analyses for a proposed approach.

Furthermore we applied the framework to analyse popular state-of-the-art gold standards for machine reading comprehension: We reveal issues with their factual correctness, show the presence of lexical cues and we observe that semantics-altering grammatical modifiers are missing in all of the investigated gold standards. 
Studying how to introduce those modifiers into gold standards and observing whether state-of-the-art MRC models are capable of performing reading comprehension on text containing them, is a future research goal.

A future line of research is to extend the framework to be able to identify the different types of exploitable cues such as question or entity typing and concrete overlap patterns. This will allow the framework to serve as an interpretable estimate of reading comprehension complexity of gold standards. Finally, investigating gold standards under this framework where MRC models outperform the human baseline (e.g. \textsc{SQuAD}) will contribute to a deeper understanding of the seemingly superb performance of deep learning approaches on them.

\section*{References}
\bibliographystyle{lrec}
\bibliography{main}

\appendix
\onecolumn
\section{Annotation Schema}
\label{sec:appendix-schema}
Here, we describe our annotation schema in greater detail. We present the respective phenomenon, give a short description and present an example that illustrates the feature. Examples for categories that occur in the analysed samples are taken directly from observed data and therefore do not represent the views, beliefs or opinions of the authors. For those categories that were not annotated in the data we construct a minimal example.
\subsection*{Supporting Fact}
We define and annotate ``Supporting fact(s)'' in line with contemporary literature as the (minimal set of) sentence(s) that is required in order to provide an answer to a given question. Other sources also call supporting facts ``evidence''.

\subsection*{Answer Type}
\paragraph{Span} We mark an answer as span if the answer is a text span from the paragraph.

\emph{Question:} Who was freed from collapsed roadway tunnel?\\
\emph{Passage:} [\ldots] The quake collapsed a roadway tunnel, temporarily trapping about 50 construction workers. [\ldots] \\
\emph{Expected Answer:} 50 construction workers.

\paragraph{Paraphrasing} We annotate an answer as paraphrasing if the expected correct answer is a paraphrase of a textual span. This can include the usage of synonyms, altering the constituency structure or changing the voice or mode.

\emph{Question:} What is the CIA known for?\\
\emph{Passage:} [\ldots] The CIA has a reputation for agility [\ldots] \\
\emph{Expected Answer:} CIA is known for agility.

\paragraph{Unanswerable}
We annotate an answer as unanswerable if the answer is not provided in the accompanying paragraph.

\emph{Question:} average daily temperature in Beaufort, SC \\
\emph{Passage:} The highest average temperature in Beaufort is June at 80.8 degrees. The coldest average temperature in Beaufort is February at 50 degrees [\ldots]. 

\paragraph{Generated} We annotate an answer as generated, if and only if it does not fall into the three previous categories. Note that neither answers that are conjunctions of previous categories (e.g. two passage spans concatenated with ``and'') nor results of concatenating passage spans or restating the question in order to formulate a full sentence (i.e. enriching it with pronomina) are counted as generated answers.

\emph{Question:} How many total points were scored in the game? \\
\emph{Passage:} [\ldots] as time expired to shock the Colts 27-24. \\
\emph{Expected Answer:} 51.

\subsection*{Quality}

\paragraph{Debatable} We annotate an answer as debatable, either if it cannot be deduced from the paragraph, if there are multiple plausible alternatives or if the answer is not specific enough. We add a note with the alternatives or a better suiting answer. 

\emph{Question:} what does carter say? \\
\emph{Passage:} [\ldots] ``From the time he began, [\ldots]'' the former president [\ldots] said in a statement.
``Jody was beside me in every decision I made [\ldots]'' \\
\emph{Expected Answer:} ``Jody was beside me in every decision I made [\ldots]'' (\emph{This is an arbitrary selection as more direct speech is attributed to Carter in the passage.})
\paragraph{Wrong} We annotate an answer as wrong, if the answer is factually wrong. Further, we denote why the answer is wrong and what the correct answer should be.

\emph{Question:} What is the cost of the project? \\
\emph{Passage:} [\ldots] At issue is the [\ldots] platform, [\ldots] that has cost taxpayers \$1.2 billion in earmarks since 2004. It is estimated to cost at least \$2.9 billion more  [\ldots]. \\
\emph{Expected Answer:} \$2.9 Billion. (\emph{The overall cost is at least \$ 4.1 Billion})


\subsection*{Linguistic Features}
We annotate occurrences of the following linguistic features in the supporting facts. On a high-level, we differentiate between syntax and lexical semantics, as well as variety and ambiguity. Naturally, features that concern question and corresponding passage context tend to fall under the variety category, while features that relate to the passage only are typically associated with the ambiguity category.
\subsubsection*{Lexical Variety}
\paragraph{Redundancy} We annotate a span as redundant, if it does not alter the factuality of the sentence. In other words the answer to the question remains the same if the span is removed (and the sentence is still grammatically correct).

\emph{Question:} When was the last time the author went to the cellars?\\
\emph{Passage:} I had not, [if I remember rightly]$_{Redundancy}$, been into [the cellars] since [my hasty search on]$_{Redundancy}$ the evening of the attack.

\paragraph{Lexical Entailment} We annotate occurrences, where it is required to navigate the semantic fields of words in order to derive the answer as lexical entailment. In other words we annotate cases, where the understanding of words' hypernymy and hyponomy relationships is necessary to arrive at the expected answer. 

\emph{Question:} What [food items]$_{LexEntailment}$ are mentioned? \\
\emph{Passage:} He couldn't find anything to eat except for [pie]$_{LexEntailment}$! Usually, Joey would eat [cereal]$_{LexEntailment}$, [fruit]$_{LexEntailment}$ (a [pear]$_{LexEntailment}$), or [oatmeal]$_{LexEntailment}$ for breakfast.
\paragraph{Dative} We annotate occurrences of variance in case of the object (i.e. from dative to using preposition) in the question and supporting facts. 

\emph{Question:} Who did Mary buy a gift for? \\
\emph{Passage:} Mary bought Jane a gift.

\paragraph{Synonym and Paraphrase} We annotate cases, where the question wording uses synonyms or paraphrases of expressions that occur in the supporting facts.

\emph{Question:} How many years longer is the life expectancy of [women]$_{Synonym}$ than [men]$_{Synonym}$? \\
\emph{Passage:} Life expectancy is [female]$_{Synonym}$ 75, [male]$_{Synonym}$ 72. 
\paragraph{Abbreviation} We annotate cases where the correct resolution of an abbreviation is required, in order to arrive at the answer.

\emph{Question:} How many [touchdowns]$_{Abbreviation}$ did the Giants score in the first half? \\
\emph{Paragraph:} [\ldots] with RB Brandon Jacobs getting a 6-yard and a 43-yard [TD]$_{Abbreviation}$ run [\ldots]
\paragraph{Symmetry, Collectivity and Core arguments}
We annotate the argument variance for the same predicate in question and passage such as argument collection for symmetric verbs or the exploitation of ergative verbs.\\
\emph{Question:} Who married John?\\
\emph{Passage:} John and Mary married.
\subsubsection*{Syntactic Variety}
\paragraph{Nominalisation} We annotate occurrences of the change in style from nominal to verbal (and vice versa) of verbs (nouns) occurring both in question and supporting facts.

\emph{Question:} What show does [the host of]$_{Nominalisation}$ The 2011 Teen Choice Awards ceremony currently star on? \\ 
\emph{Passage:} The 2011 Teen Choice Awards ceremony, [hosted by]$_{Nominalisation}$ Kaley Cuoco, aired live on August 7, 2011 at 8/7c on Fox.

\paragraph{Genitives} We annotate cases where possession of an object is expressed by using the genitive form (``'s'') in question and differently (e.g. using the preposition ``of'') in the supporting facts (and vice versa).

\emph{Question:} Who used Mary's computer? \\
\emph{Passage:} John's computer was broken, so he went to Mary's office where he used the computer of Mary.

\paragraph{Voice} We annotate occurrences of the change in voice from active to passive (and vice versa) of verbs shared by question and supporting facts.

\emph{Question:} Where does Mike Leach currently [coach at]$_{Voice}$? \\
\emph{Passage:} [The 2012 Washington State Cougars football team] was [coached]$_{Voice}$ by by first-year head coach Mike Leach [...].
\subsubsection*{Lexical Ambiguity}
\paragraph{Restrictivity} We annotate cases where restrictive modifiers need to be resolved in order to arrive at the expected answers. Restrictive modifiers -- opposed to redundancy -- are modifiers that change the meaning of a sentence by providing additional details. 

\emph{Question:} How many dogs are in the room? \\
\emph{Passage:} There are 5 dogs in the room. Three of them are brown. All the [brown]$_{Restrictivity}$ dogs leave the room.

\paragraph{Factivity} We annotate cases where modifiers -- such as verbs -- change the factivity of a statement.

\emph{Question:} When did it rain the last time? \\
\emph{Passage:} Upon reading the news, I realise that it rained two days ago. I believe it rained yesterday. \\ 
\emph{Expected Answer:} two days ago

\paragraph{Coreference} We annotate cases where intra- or inter-sentence coreference and anaphora need to be resolved in order to retrieve the expected answer. 

\emph{Question:} What is the name of the psychologist who is known as the originator of social learning theory?\\
\emph{Passage:} Albert Bandura OC (born December 4, 1925) is a psychologist who is the David Starr Jordan Professor Emeritus of Social Science in Psychology at Stanford University. [\ldots] He is known as the originator of social learning theory and the theoretical construct of self-efficacy, and is also responsible for the influential 1961 Bobo doll experiment.

\paragraph{Ellipsis/Implicit} We annotate cases where required information is not explicitly expressed in the passage.

\emph{Question:} How many years after producing Happy Days did Beckett produce Rockaby?\\
\emph{Passage:} [Beckett] produced works [\ldots], including [...], Happy Days [(1961)]$_{Implicit}$, and Rockaby [(1981)]$_{Implicit}$. \emph{(The date in brackets indicates the publication date implicitly.)}

\subsubsection*{Syntactic Ambiguity}
\paragraph{Preposition} We annotate occurrences of ambiguous prepositions that might obscure the reasoning process if resolved incorrectly.

\emph{Question:} What tool do you eat spaghetti with? \\
\emph{Passage:} Let's talk about forks. You use them to eat spaghetti with meatballs.

\paragraph{Listing} We define listing as the case where multiple arguments belonging to the same predicate are collected with conjunctions or disjunctions (i.e. ``and'' or ``or''). We annotate occurrences of listings where the resolution of such collections and mapping to the correct predicate is required in order to obtain the information required to answer the question.  

\emph{Passage:} [She is also known for her roles]$_{Predicate}$ [as White House aide Amanda Tanner in the first season of ABC's "Scandal"]$_{Argument}$ [and]$_{Listing}$ [as attorney Bonnie Winterbottom in ABC's "How to Get Away with Murder"]$_{Argument}$.

\paragraph{Coordination Scope} We annotate cases where the scope of a coordination may be interpreted differently and thus lead to a different answer than the expected one.
\emph{Question:} Where did I put the marbles? \\
\emph{Passage:} I put the marbles in the box and the bowl on the table. \emph{Depending on the interpretation, the marbles were either put both in the box and in the bowl that was on the table, or the marbles were put in the box and the bowl was put on the table.}

\paragraph{Relative clause, adverbial phrase and apposition} We annotate cases that require the correct resolution of relative pronomina, adverbial phrases or appositions in order to answer a question correctly.

\emph{Question:} Jos\'e Saramago and Ivo Andri\'c were recipients of what award in Literature? \\
\emph{Passage:} Ivo Andri\'c [\ldots] was a Yugoslav novelist, poet and short story writer [who]$_{Relative}$ won the Nobel Prize in Literature in 1961.

\subsection*{Required Reasoning}
\subsubsection*{Operational Reasoning}
We annotate occurrences of the arithmetic operations described below. Operational reasoning is a type of abstract reasoning, which means that we do not annotate passages that explicitly state the information required to answer the question, even if the question's wording might indicate it. For example, we don't regard the reasoning in the question ``How many touchdowns did the Giants score in the first half?'' as operational (counting) if the passage states ``The Giants scored 2 touchdowns in the first half.''
\paragraph{Bridge} We annotate cases where information to answer the question needs to be gathered from multiple supporting facts, ``bridged'' by commonly mentioned entities, concepts or events. This phenomenon is also known as ``Multi-hop reasoning'' in literature.

\emph{Question:} What show does the host of The 2011 Teen Choice Awards ceremony currently star on? \\
\emph{Passage:} [\ldots] The 2011 Teen Choice Awards ceremony, hosted by [Kaley Cuoco]$_{Entity}$, aired live on August 7, 2011 at 8/7c on Fox. [\ldots] [Kaley Christine Cuoco]$_{Entity}$ is an American actress. Since 2007, she has starred as Penny on the CBS sitcom "The Big Bang Theory", for which she has received Satellite, Critics' Choice, and People's Choice Awards.

\paragraph{Comparison} We annotate questions where entities, concepts or events needs to be compared with regard to their properties in order to answer a question.

\emph{Question:} What year was the alphabetically first writer of Fairytale of New York born? \\
\emph{Passage:} "Fairytale of New York" is a song written by Jem Finer and Shane MacGowan [\ldots].
\paragraph{Constraint Satisfaction} Similar to the Join category, we annotate instances that require the retrieval of entities, concepts or events which additionally satisfy a specified constraint.

\emph{Question:} Which Australian singer-songwriter wrote Cold Hard Bitch? \\
\emph{Passage:} [``Cold Hard Bitch''] was released in March 2004 and was written by band-members Chris Cester, Nic Cester, and Cameron Muncey. [\ldots] Nicholas John "Nic" Cester is an Australian singer-songwriter and guitarist [\ldots].
\paragraph{Intersection} Similar to the Comparison category, we annotate cases where properties of entities, concepts or events need need to be reduced to a minimal common set. 

\emph{Question:} Jos\'e Saramago and Ivo Andri\'c were recipients of what award in Literature? 

\subsubsection*{Arithmetic Reasoning}
We annotate occurrences of the arithmetic operations described below. Similarly to operational reasoning, arithmetic reasoning is a type of abstract reasoning, so we annotate it analogously. An example for \emph{non-arithmetic} reasoning is, if the question states ``How many total points were scored in the game?'' and the passage expresses the required information similarly to ``There were a total of 51 points scored in the game.''
\paragraph{Substraction}

\emph{Question:} How many points were the Giants behind the Dolphins at the start of the 4th quarter? \\
\emph{Passage:} New York was down 17-10 behind two rushing touchdowns.

\paragraph{Addition}

\emph{Question:} How many total points were scored in the game? \\
\emph{Passage:} [\ldots] Kris Brown kicked the winning 48-yard field goal as time expired to shock the Colts 27-24.
\paragraph{Ordering} We annotate questions with this category, if it requires the comparison of (at least) two numerical values (and potentially a selection based on this comparison) to produce the expected answer.

\emph{Question:} What happened second: Peace of Paris or appointed governor of Artois? \\
\emph{Passage:} He [\ldots] retired from active military service when the war ended in 1763 with the Peace of Paris. He was appointed governor of Artois in 1765.


\paragraph{Count} We annotate questions that require the explicit enumeration of events, concepts, facts or entities.

\emph{Question:} How many touchdowns did the Giants score in the first half?\\
\emph{Passage:} In the second quarter, the Giants took the lead with RB Brandon Jacobs getting a 6-yard and a 43-yard TD run [\ldots].

\paragraph{Other} We annotate any other arithmetic operation that does not fall into any of the above categories with this label. 

\emph{Question:} How many points did the Ravens score on average? \\
\emph{Passage:} Baltimore managed to beat the Jets 10-9 on the 2010 opener [\ldots]. The Ravens rebounded [\ldots], beating Cleveland 24-17 in Week 3 and then Pittsburgh 17-14 in Week 4. [\ldots] Next, the Ravens hosted Miami and won 26-10, breaking that teams 4-0 road streak.
\subsubsection*{Linguistic Reasoning}
\paragraph{Negations} We annotate cases where the information in the passage needs to be negated in order to conclude the correct answer.

\emph{Question:} How many percent are not Marriage couples living together? \\
\emph{Passage:} [\ldots] 46.28\% were Marriage living together. [\ldots]
\paragraph{Conjunctions and Disjunctions} We annotate occurrences, where in order to conclude the answer logical conjunction or disjunction needs to be resolved.

\emph{Question:} Is dad in the living room?\\
\emph{Passage:} Dad is either in the kitchen or in the living room.
\paragraph{Conditionals} We annotate cases where the the expected answer is guarded by a condition. In order to arrive at the answer, the inspection whether the condition holds is required.

\emph{Question:} How many eggs did I buy? \\
\emph{Passage:} I am going to buy eggs. If you want some, too, I will buy 6, if not I will buy 3. You didn't want any.
\paragraph{Quantification} We annotate occurrences, where it is required to understand the concept of quantification (existential and universal) in order to determine the correct answer.

\emph{Question:} How many presents did Susan receive? \\
\emph{Passage:} On the day of the party, all five friends showed up. [Each friend]$_{Quantification}$ had a present for Susan.

\subsubsection*{Other types of reasoning}
\paragraph{Temporal} We annotate cases, where understanding about the succession is required in order to derive an answer. Similar to arithmetic and operational reasoning, we do not annotate questions where the required information is expressed explicitly in the passage.

\emph{Question:} Where is the ball? \\
\emph{Passage:} I take the ball. I go to the kitchen after going to the living room. I drop the ball. I go to the garden.
\paragraph{Spatial} Similarly to temporal, we annotate cases where understanding about directions, environment and spatiality is required in order to arrive at the correct conclusion.

\emph{Question:} What is the 2010 population of the city 2.1 miles southwest of Marietta Air Force Station?
\emph{Passage:} [Marietta Air Force Station] is located 2.1 mi northeast of Smyrna, Georgia.
\paragraph{Causal}
We annotate occurrences where causal (i.e. cause-effect) reasoning between events, entities or concepts is required to correctly answer a question. We do not annotate questions as causal, if passages explicitly reveal the relationship in a ``effect because cause'' manner. For example we don't annotate ``Why do men have a hands off policy when it comes to black women's hair?'' as causal, even if the wording indicates it, because the corresponding passage immideately reveals the relationship by stating ``Because women spend so much time and money on their hair, Rock says men are forced to adopt a hands-off policy.''.

\emph{Question:} Why did Sam stop Mom from making four sandwich? \\
\emph{Passage:} [\ldots] There are three of us, so we need three sandwiches. [\ldots]

\paragraph{By Exclusion}
We annotate occurrences (in the multiple-choice setting) where there is not enough information present to directly determine the expected answer, and the expected answer can only be assumed by excluding alternatives.

\emph{Question:} Calls for a withdrawal of investment in Israel have also intensified because of its continuing occupation of @placeholder territories -- something which is illegal under international law.\\
\emph{Answer Choices} 
Benjamin Netanyahu,
Paris, 
[Palestinian]$_{Answer}$, 
French,
Israeli, 
Partner's, 
West Bank, 
Telecoms,
Orange
\paragraph{Information Retrieval} We collect cases that don't fall under any of the described categories and where the answer can be directly retrieved from the passage under this category. 

\emph{Question:} Officers were fatally shot where? \\
\emph{Passage:} The Lakewood police officers [...] were fatally shot November 29 [in a coffee shop near Lakewood]$_{Answer}$.
\subsection*{Knowledge} We recognise passages that do not contain the required information in order to answer a question as expected. These non self sufficient passages require models to incorporate some form of \emph{external knowledge}. We distinguish between factual and common sense knowledge.
\subsubsection*{Factual} We annotate the dependence on factual knowledge -- knowledge that can clearly be stated as a set facts -- from the domains listed below.
\paragraph{Cultural/Historic} 
\emph{Question:} What are the details of the second plot on Alexander's life in the Central Asian campaign?\\
\emph{Passage:} Later, in the Central Asian campaign, a second plot against his life was revealed, this one instigated by his own royal pages. His official historian, Callisthenes of Olynthus, was implicated in the plot; however, historians have yet to reach a consensus regarding this involvement.\\
\emph{Expected Answer:} Unsuccessful
\paragraph{Geographical/Political}
\emph{Question:} Calls for a withdrawal of investment in Israel have also intensified because of its continuing occupation of @placeholder territories -- something which is illegal under international law. \\
\emph{Passage:} [\ldots]  But Israel lashed out at the decision, which appeared to be related to Partner's operations in the occupied West Bank. [\ldots] \\
\emph{Expected Answer:} Palestinian
\paragraph{Legal} 
\emph{Question:} [\ldots] in part due to @placeholder -- the 1972 law that increased opportunities for women in high school and college athletics -- and a series of court decisions. \\
\emph{Passage:} [\ldots] Title IX helped open opportunity to women too; Olympic hopeful Marlen Exparza one example. [\ldots]
\emph{Expected Answer:} Title IX
\paragraph{Technical/Scientific}
\emph{Question:} What are some renewable resources? \\
\emph{Passage:} [\ldots] \emph{plants are not mentioned in the passage} [\ldots] \\
\emph{Expected Answer:} Fish, plants
\paragraph{Other Domain Specific}
\emph{Question:} Which position scored the shortest touchdown of the game? \\
\emph{Passage:} [\ldots] However, Denver continued to pound away as RB Cecil Sapp got a 4-yard TD run, while kicker Jason Elam got a 23-yard field goal. [\ldots] \\
\emph{Expected Answer:} RB
\subsubsection*{Intuitive} We annotate the requirement of intuitive knowledge in order to answer a question common sense knowledge. Opposed to factual knowledge, it is hard to express as a set of facts.

\emph{Question:} Why would Alexander have to declare an heir on his deathbed? \\
\emph{Passage:}  According to Diodorus, Alexander's companions asked him on his deathbed to whom he bequeathed his kingdom; his laconic reply was "toi kratistoi"--"to the strongest". \\
\emph{Expected Answer:} So that people know who to follow.

\section{Detailed annotation results}
\label{sec:appendix-results}
Here, we report all our annotations in detail, with absolute and relative numbers. Note, that numbers from sub-categories do not necessarily add up to the higher level category, because an example might contain features from the same higher-level category. (for example if an example requires both Bridge and Constraint type of reasoning, it will still count as a single example towards the \emph{Operations} counter).

\begin{table}[!hptb]
\centering
\begin{tabularx}{1\columnwidth}{ X | c | c | c | c | c | c | c | c | c | c | c | c }
\multirow{2}{*}{ } & \multicolumn{2}{c|}{\textsc{MSMarco}} & \multicolumn{2}{c|}{\textsc{HotpotQA}} & \multicolumn{2}{c|}{\textsc{ReCoRd}} & \multicolumn{2}{c|}{\textsc{MultiRC}} & \multicolumn{2}{c|}{\textsc{NewsQA}} & \multicolumn{2}{c}{\textsc{DROP}} \\
& abs. & rel. &  abs. & rel. & abs. & rel. & abs. & rel. & abs. & rel. & abs. & rel. \\
\hline
\hline
\textbf{Answer} & 50 & 100.0 & 50 & 100.0 & 50 & 100.0 & 50 & 100.0 & 50 & 100.0 & 50 & 100.0\\
\hline
Span & 25 & 50.0 & 49 & 98.0 & 50 & 100.0 & 36 & 72.0 & 38 & 76.0 & 20 & 40.0\\
Paraphrasing & 4 & 8.0 & 0 & 0.0 & 0 & 0.0 & 24 & 48.0 & 0 & 0.0 & 0 & 0.0\\
Unanswerable & 20 & 40.0 & 0 & 0.0 & 0 & 0.0 & 0 & 0.0 & 12 & 24.0 & 0 & 0.0\\
Abstraction & 1 & 2.0 & 1 & 2.0 & 0 & 0.0 & 12 & 24.0 & 0 & 0.0 & 31 & 62.0\\
\hline
\end{tabularx}
\caption{Detailed Answer Type results. We calculate percentages relative to the number of examples in the sample.}
\label{tab:answers}
\end{table}
\begin{table}[!hptb]
\centering
\begin{tabularx}{1\columnwidth}{ X | c | c | c | c | c | c | c | c | c | c | c | c }
\multirow{2}{*}{ } & \multicolumn{2}{c|}{\textsc{MSMarco}} & \multicolumn{2}{c|}{\textsc{HotpotQA}} & \multicolumn{2}{c|}{\textsc{ReCoRd}} & \multicolumn{2}{c|}{\textsc{MultiRC}} & \multicolumn{2}{c|}{\textsc{NewsQA}} & \multicolumn{2}{c}{\textsc{DROP}} \\
& abs. & rel. &  abs. & rel. & abs. & rel. & abs. & rel. & abs. & rel. & abs. & rel. \\
\hline
\hline
\textbf{Factual Correctness} & 23 & 46.0 & 13 & 26.0 & 4 & 8.0 & 19 & 38.0 & 21 & 42.0 & 5 & 10.0\\
Debatable & 17 & 34.0 & 12 & 24.0 & 4 & 8.0 & 14 & 28.0 & 16 & 32.0 & 5 & 10.0\\
\hline
Arbitrary Selection & 9 & 18.0 & 2 & 4.0 & 0 & 0.0 & 0 & 0.0 & 5 & 10.0 & 1 & 2.0\\
Arbitrary Precision & 3 & 6.0 & 5 & 10 & 1 & 2.0 & 4 & 8.0 & 7 & 14.0 & 2 & 4.0\\
Conjunction or Isolated & 0 & 0.0 & 0 & 0 & 0 & 0.0 & 5 & 10.0 & 0 & 0.0 & 0 & 0.0\\
Other   & 5 & 10.0 & 5 & 10 & 3 & 6.0 & 5 & 10.0 & 4 & 8.0 & 2 & 4.0\\
\hline
Wrong & 6 & 12.0 & 1 & 2.0 & 0 & 0.0 & 5 & 10.0 & 5 & 10.0 & 0 & 0.0\\
\hline
\end{tabularx}
\caption{Detailed results for the annotation of factual correctness.}
\label{tab:correctness}
\end{table}
\begin{table}[!hptp]
\centering
\begin{tabularx}{1\columnwidth}{ X | c | c | c | c | c | c | c | c | c | c | c | c }
\multirow{2}{*}{ } & \multicolumn{2}{c|}{\textsc{MSMarco}} & \multicolumn{2}{c|}{\textsc{HotpotQA}} & \multicolumn{2}{c|}{\textsc{ReCoRd}} & \multicolumn{2}{c|}{\textsc{MultiRC}} & \multicolumn{2}{c|}{\textsc{NewsQA}} & \multicolumn{2}{c}{\textsc{DROP}} \\
& abs. & rel. &  abs. & rel. & abs. & rel. & abs. & rel. & abs. & rel. & abs. & rel. \\
\hline
\hline
\textbf{Knowledge} & 3 & 10.0 & 8 & 16.0 & 19 & 38.0 & 11 & 22.0 & 6 & 15.8 & 20 & 40.0\\
\hline
\textit{World} & 0 & 0.0 & 3 & 6.0 & 12 & 24.0 & 3 & 6.0 & 1 & 2.6 & 6 & 12.0\\
\hline
Cultural & 0 & 0.0 & 1 & 2.0 & 3 & 6.0 & 1 & 2.0 & 0 & 0.0 & 0 & 0.0\\
Geographical & 0 & 0.0 & 0 & 0.0 & 2 & 4.0 & 0 & 0.0 & 1 & 2.6 & 0 & 0.0\\
Legal & 0 & 0.0 & 0 & 0.0 & 2 & 4.0 & 0 & 0.0 & 0 & 0.0 & 0 & 0.0\\
Political & 0 & 0.0 & 1 & 2.0 & 2 & 4.0 & 0 & 0.0 & 0 & 0.0 & 1 & 2.0\\
Technical & 0 & 0.0 & 0 & 0.0 & 1 & 2.0 & 2 & 4.0 & 0 & 0.0 & 0 & 0.0\\
DomainSpecific & 0 & 0.0 & 1 & 2.0 & 2 & 4.0 & 0 & 0.0 & 0 & 0.0 & 5 & 10.0\\
\hline
Intuitive & 3 & 10.0 & 5 & 10.0 & 9 & 18.0 & 8 & 16.0 & 5 & 13.2 & 14 & 28.0\\

\end{tabularx}
\caption{Detailed results for the annotation of factual correctness. We calculate percentages relative to the number of examples that were annotated to be not unanswerable.}
\label{tab:knowledge}
\end{table}

\begin{table}[!hptb]
\centering
\begin{tabularx}{1\columnwidth}{ X | c | c | c | c | c | c | c | c | c | c | c | c }
\multirow{2}{*}{ } & \multicolumn{2}{c|}{\textsc{MSMarco}} & \multicolumn{2}{c|}{\textsc{HotpotQA}} & \multicolumn{2}{c|}{\textsc{ReCoRd}} & \multicolumn{2}{c|}{\textsc{MultiRC}} & \multicolumn{2}{c|}{\textsc{NewsQA}} & \multicolumn{2}{c}{\textsc{DROP}} \\
& abs. & rel. &  abs. & rel. & abs. & rel. & abs. & rel. & abs. & rel. & abs. & rel. \\
\hline
\hline
\textbf{Reasoning} & 30 & 1.0 & 50 & 1.0 & 50 & 1.0 & 50 & 1.0 & 38 & 1.0 & 50 & 1.0\\
\hline
\textit{Mathematics} & 0 & 0.0 & 3 & 6.0 & 0 & 0.0 & 1 & 2.0 & 0 & 0.0 & 34 & 68.0\\
\hline
Subtraction & 0 & 0.0 & 0 & 0.0 & 0 & 0.0 & 1 & 2.0 & 0 & 0.0 & 20 & 40.0\\
Addition & 0 & 0.0 & 0 & 0.0 & 0 & 0.0 & 0 & 0.0 & 0 & 0.0 & 2 & 4.0\\
Ordering & 0 & 0.0 & 3 & 6.0 & 0 & 0.0 & 0 & 0.0 & 0 & 0.0 & 11 & 22.0\\
OtherArithmethic & 0 & 0.0 & 0 & 0.0 & 0 & 0.0 & 0 & 0.0 & 0 & 0.0 & 2 & 4.0\\
\hline
\textit{Linguistics} & 2 & 6.7 & 0 & 0.0 & 2 & 4.0 & 7 & 14.0 & 0 & 0.0 & 2 & 4.0\\
\hline
Negation & 0 & 0.0 & 0 & 0.0 & 2 & 4.0 & 1 & 2.0 & 0 & 0.0 & 2 & 4.0\\
Con-/Disjunction & 0 & 0.0 & 0 & 0.0 & 0 & 0.0 & 1 & 2.0 & 0 & 0.0 & 0 & 0.0\\
Conditionals & 0 & 0.0 & 0 & 0.0 & 0 & 0.0 & 0 & 0.0 & 0 & 0.0 & 0 & 0.0\\
Monotonicity & 0 & 0.0 & 0 & 0.0 & 0 & 0.0 & 0 & 0.0 & 0 & 0.0 & 0 & 0.0\\
Quantifiers & 0 & 0.0 & 0 & 0.0 & 0 & 0.0 & 0 & 0.0 & 0 & 0.0 & 0 & 0.0\\
Exists & 2 & 6.7 & 0 & 0.0 & 0 & 0.0 & 4 & 8.0 & 0 & 0.0 & 0 & 0.0\\
ForAll & 0 & 0.0 & 0 & 0.0 & 0 & 0.0 & 1 & 2.0 & 0 & 0.0 & 0 & 0.0\\
\hline
\textit{Operations} & 2 & 6.7 & 36 & 72.0 & 0 & 0.0 & 1 & 2.0 & 2 & 5.3 & 8 & 16.0\\
\hline
Join & 1 & 3.3 & 23 & 46.0 & 0 & 0.0 & 0 & 0.0 & 0 & 0.0 & 0 & 0.0\\
Comparison & 1 & 3.3 & 2 & 4.0 & 0 & 0.0 & 0 & 0.0 & 0 & 0.0 & 0 & 0.0\\
Count & 0 & 0.0 & 0 & 0.0 & 0 & 0.0 & 0 & 0.0 & 0 & 0.0 & 7 & 14.0\\
Constraint & 0 & 0.0 & 11 & 22.0 & 0 & 0.0 & 1 & 2.0 & 2 & 5.3 & 6 & 12.0\\
Intersection & 0 & 0.0 & 4 & 8.0 & 0 & 0.0 & 0 & 0.0 & 0 & 0.0 & 0 & 0.0\\
\hline
Temporal & 0 & 0.0 & 0 & 0.0 & 0 & 0.0 & 0 & 0.0 & 0 & 0.0 & 0 & 0.0\\
Spatial & 0 & 0.0 & 1 & 2.0 & 0 & 0.0 & 0 & 0.0 & 0 & 0.0 & 0 & 0.0\\
Causal & 0 & 0.0 & 0 & 0.0 & 2 & 4.0 & 15 & 30.0 & 0 & 0.0 & 0 & 0.0\\
ByExclusion & 0 & 0.0 & 0 & 0.0 & 17 & 34.0 & 1 & 2.0 & 0 & 0.0 & 0 & 0.0\\
Retrieval & 26 & 86.7 & 13 & 26.0 & 31 & 62.0 & 30 & 60.0 & 38 & 100.0 & 9 & 18.0\\

\hline
\end{tabularx}
\caption{Detailed reasoning results. We calculate percentages relative to the number of examples that are not unanswerable, i.e. require reasoning to obtain the answer according to our definition.}
\label{tab:reasoning}
\end{table}

\begin{table}[!hptb]
\centering
\begin{tabularx}{1\columnwidth}{ X | c | c | c | c | c | c | c | c | c | c | c | c }
\multirow{2}{*}{ } & \multicolumn{2}{c|}{\textsc{MSMarco}} & \multicolumn{2}{c|}{\textsc{HotpotQA}} & \multicolumn{2}{c|}{\textsc{ReCoRd}} & \multicolumn{2}{c|}{\textsc{MultiRC}} & \multicolumn{2}{c|}{\textsc{NewsQA}} & \multicolumn{2}{c}{\textsc{DROP}} \\
& abs. & rel. &  abs. & rel. & abs. & rel. & abs. & rel. & abs. & rel. & abs. & rel. \\
\hline
\hline
\textbf{LinguisticComplexity} & 18 & 60.0 & 49 & 98.0 & 42 & 97.7 & 43 & 87.8 & 34 & 89.5 & 46 & 92.0\\
\hline
\emph{Lexical Variety} & 14 & 46.7 & 44 & 88.0 & 36 & 83.7 & 35 & 71.4 & 30 & 78.9 & 42 & 84.0\\
\hline
Redundancy & 12 & 40.0 & 38 & 76.0 & 19 & 44.2 & 31 & 63.3 & 27 & 71.1 & 30 & 60.0\\
Lex Entailment  & 0 & 0.0 & 1 & 2.0 & 1 & 2.3 & 2 & 4.1 & 0 & 0.0 & 0 & 0.0\\
Dative & 0 & 0.0 & 0 & 0.0 & 0 & 0.0 & 0 & 0.0 & 0 & 0.0 & 0 & 0.0\\
Synonym & 7 & 23.3 & 7 & 14.0 & 25 & 58.1 & 11 & 22.4 & 15 & 39.5 & 12 & 24.0\\
Abbreviation & 2 & 6.7 & 4 & 8.0 & 1 & 2.3 & 1 & 2.0 & 0 & 0.0 & 7 & 14.0\\
Symmetry & 0 & 0.0 & 0 & 0.0 & 0 & 0.0 & 0 & 0.0 & 0 & 0.0 & 0 & 0.0\\
\hline
\emph{Syntactic Variety} & 2 & 6.7 & 10 & 20.0 & 2 & 4.7 & 2 & 4.1 & 1 & 2.6 & 4 & 8.0\\
\hline
Nominalisation & 0 & 0.0 & 6 & 12.0 & 0 & 0.0 & 1 & 2.0 & 0 & 0.0 & 2 & 4.0\\
Genitive & 0 & 0.0 & 0 & 0.0 & 0 & 0.0 & 0 & 0.0 & 0 & 0.0 & 0 & 0.0\\
Voice & 2 & 6.7 & 4 & 8.0 & 2 & 4.7 & 1 & 2.0 & 1 & 2.6 & 2 & 4.0\\
\hline
\emph{Lexical Ambiguity} & 7 & 23.3 & 32 & 64.0 & 26 & 60.5 & 34 & 69.4 & 11 & 28.9 & 7 & 14.0\\
\hline
Coreference & 7 & 23.3 & 32 & 64.0 & 26 & 60.5 & 34 & 69.4 & 11 & 28.9 & 7 & 14.0\\
Restrictivity & 0 & 0.0 & 0 & 0.0 & 0 & 0.0 & 0 & 0.0 & 0 & 0.0 & 0 & 0.0\\
Factivity  & 0 & 0.0 & 0 & 0.0 & 0 & 0.0 & 1 & 2.0 & 0 & 0.0 & 0 & 0.0\\
\hline
\emph{Syntactic Ambiguity} & 2 & 6.7 & 22 & 44.0 & 6 & 14.0 & 7 & 14.3 & 9 & 23.7 & 9 & 18.0\\
\hline
Preposition & 0 & 0.0 & 1 & 2.0 & 0 & 0.0 & 0 & 0.0 & 0 & 0.0 & 0 & 0.0\\
Ellipse/Implicit & 2 & 6.7 & 3 & 6.0 & 3 & 7.0 & 3 & 6.1 & 1 & 2.6 & 8 & 16.0\\
Listing & 0 & 0.0 & 16 & 32.0 & 5 & 11.6 & 6 & 12.2 & 1 & 2.6 & 13 & 26.0\\
Scope & 0 & 0.0 & 0 & 0.0 & 0 & 0.0 & 0 & 0.0 & 0 & 0.0 & 0 & 0.0\\
Relative & 0 & 0.0 & 20 & 40.0 & 3 & 7.0 & 4 & 8.2 & 8 & 21.1 & 3 & 6.0\\
\hline
\end{tabularx}
\caption{Detailed linguistic feature results. We calculate percentages relative to the number of examples that were annotated to contain supporting facts.}
\label{tab:linguistic}
\end{table}

\section{Description of selected gold standards}
\label{sec:appendix-datasets}
\paragraph{\scshape MSMarco} was created by sampling real user queries from the log of a search engine and presenting the search results to experts in order to select relevant passages. Those passages were then shown to crowd workers in order to generate a free-form answer that answers the question or mark if the question is not answerable from the given context. While the released dataset can be used for a plethora of tasks we focus on the MRC aspect where the task is to predict an expected answer (if existent), given a question and ten passages that are extracted from web documents.

\paragraph{\scshape HotpotQA} is a dataset and benchmark that focuses on ``multi-hop'' reasoning, i.e. information integration from different sources. To that end the authors build a graph from a  where nodes represent first paragraphs of Wikipedia articles and edges represent the hyperlinks between them. They present pairs of adjacent articles from that graph or from lists of similar entities to crowd-workers and request them to formulate questions based on the information from both articles and also mark the supporting facts. The benchmark comes in two settings: We focus on the \emph{distractor} setting, where question and answer are accompanied by a context comprised of the two answer source articles and eight similar articles retrieved by a information retrieval system. 

\paragraph{\scshape ReCoRd} is automatically generated from news articles, as an attempt to reduce bias introduced by human annotators. The benchmark entries are comprised of an abstractive summary of a news article and a close-style query. The query is generated by sampling from a set of sentences of the full article that share any entity mention with the abstract and by removing that entity. In a final step, the machine-generated examples were presented to crowd workers to remove noisy data. The task is to predict the correct entity given the Cloze-style query and the summary.

\paragraph{\scshape MultiRC} features passages from various domains such as news, (children) stories, or textbooks. Those passages are presented to crowd workers that are required to perform the following four tasks: \emph{(i)} produce questions based multiple sentences from a given paragraph, \emph{(ii)} ensure that a question cannot be answered from any single sentence, \emph{(iii)} generate a variable number of correct and incorrect answers and \emph{(iv)} verify the correctness of produced question and answers. This results in a benchmark where the task is to predict a variable number of correct natural language answers from a variable number of choices, given a paragraph and a question.  

\paragraph{\scshape NewsQA} is generated from news articles,  similarly to \textsc{ReCoRd}, however by employing a crowd-sourcing pipeline. Question producing crowd workers were asked to formulate questions given headlines and bullet-point summaries. A different set of answer producing crowd workers was tasked to highlight the answer from the article full text or mark a question as unanswerable. A third set of crowd workers selected the best answer per question. The resulting task is, given a question and a news article to predict a span-based answer from the article. 

\paragraph{\scshape DROP} introduces explicit discrete operations to the realm of machine reading comprehension as models are expected to solve simple arithmetic tasks (such as addition, comparison, counting, etc) in order to produce the correct answer. The authors collected passages with a high density of numbers, NFL game summaries and history articles and presented them to crowd workers in order to produce questions and answers that fall in one of the aforementioned categories. A submission was only accepted, if the question was not answered correctly by a pre-trained model that was employed on-line during the annotation process, acting as an adversary. The final task is, given question and a passage to predict an answer, either as a single or multiple spans from the passage or question, generate an integer or a date.
\section{Introductory Example}
\label{sec:appendix-example}
\begin{figure}[h]
 \centering
\centering
\begin{tabularx}{1\columnwidth}{| X |}
\hline
\textbf{Passage 1: Marietta Air Force Station}\\
\emph{Marietta Air Force Station (ADC ID: M-111, NORAD ID: Z-111) is a 
closed United States Air Force General Surveillance Radar station. It is located 2.1 mi northeast of Smyrna, Georgia. It was closed in 1968.}\\
\hline
\textbf{Passage 2: Smyrna, Georgia}\\
\emph{Smyrna is a city northwest of the neighborhoods of Atlanta. It is 
in the inner ring of the Atlanta Metropolitan Area. As of the {\color{blue}2010} census, the city had a population of \textbf{51,271}. The U.S. Census Bureau estimated the population in 2013 to be 53,438. It is included in the Atlanta-Sandy Springs-Roswell MSA, which is included in the Atlanta-Athens-Clarke-Sandy Springs CSA. Smyrna grew by 28\% between the years 2000 and 2012. It is historically one of the fastest growing cities in the State of Georgia, and one of the most densely populated cities in the metro area.}\\
\hline
\textbf{Passage 3: RAF Warmwell}\\
\emph{RAF Warmwell is a former Royal Air Force station near Warmwell in Dorset, England from 1937 to 1946, located about 5 miles east-southeast of Dorchester; 100 miles southwest of London.}\\
\hline
\textbf{Passage 4: Camp Pedricktown radar station}\\
\emph{ The Camp Pedricktown Air Defense Base was a Cold War Missile Master installation with an Army Air Defense Command Post, and associated search, height finder, and identification friend or foe radars. The station's radars were subsequently replaced with radars at Gibbsboro Air Force Station 15 miles away. The obsolete Martin AN/FSG-1 Antiaircraft Defense System,a 1957-vintage vacuum tube computer, was removed after command of the defense area was transferred to the command post at Highlands Air Force Station near New York City. The Highlands AFS command post controlled the combined New York-Philadelphia Defense Area.}\\
\hline
\textbf{Passage 5: 410th Bombardment Squadron}\\
\emph{The 410th Bombardment Squadron is an inactive United States Air Force unit. It was last assigned to the 94th Bombardment Group. It was inactivated at Marietta Air Force Base, Georgia on 20 March 1951.}\\
\hline
\textbf{Passage 6: RAF Cottesmore}\\
\emph{ Royal Air Force Station Cottesmore or more simply RAF Cottesmore is a former Royal Air Force station in Rutland, England, situated between Cottesmore and Market Overton. The station housed all the operational Harrier GR9 squadrons in the Royal Air Force, and No. 122 Expeditionary Air Wing. On 15 December 2009 it was announced that the station would close in 2013 as part of defence spending cuts, along with the retirement of the Harrier GR9 and the disbandment of Joint Force Harrier. However the formal closing ceremony took place on 31 March 2011 with the airfield becoming a satellite to RAF Wittering until March 2012.}\\
\hline
\textbf{Stramshall}\\
\emph{ Stramshall is a village within the civil parish of Uttoxeter Rural in the county of Staffordshire, England. The village is 2.1 miles north of the town of Uttoxeter, 16.3 miles north east of Stafford and 143 miles north west of London. The village lies 0.8 miles north of the A50 that links Warrington to Leicester. The nearest railway station is at Uttoxeter for the Crewe to Derby line. The nearest airport is East Midlands Airport.}\\
\hline
\textbf{Topsham Air Force Station}\\
\emph{ Topsham Air Force Station is a closed United States Air Force station. It is located 2.1 mi north of Brunswick, Maine. It was closed in 1969}\\
\hline
\textbf{302d Air Division}\\
\emph{The 302d Air Division is an inactive United States Air Force Division. Its last assignment was with Fourteenth Air Force at Marietta Air Force Base, Georgia, where it was inactivated on 27 June 1949.}\\
\hline
\textbf{Eldorado Air Force Station}\\
\emph{ Eldorado Air Force Station located 35 miles south of San Angelo, Texas was one of the four unique AN/FPS-115 PAVE PAWS, early-warning phased-array radar systems. The 8th Space Warning Squadron, 21st Space Wing, Air Force Space Command operated at Eldorado Air Force Station.}\\
\hline
\hline
\textbf{Question:} \emph{What is the {\color{blue}2010} population of the city 2.1 miles southwest of Marietta Air Force Station?} \\
\textbf{Expected Answer} \emph{51,271} \\
\hline
\end{tabularx}
\caption{Full example from the Introduction.}
\label{fig:exploitable-example-2}
\end{figure}
\end{document}